%% file: main.tex
\documentclass[sigconf, balance=false]{acmart}
\usepackage{popets}

\input{math_commands.tex}

\usepackage{wrapfig}
\usepackage{tabularx}
\usepackage{enumitem} %
\usepackage{algorithm}
\usepackage{algorithmic}
\usepackage{listings} %

\setcopyright{popets}
\copyrightyear{YYYY}

\acmYear{YYYY}
\acmVolume{YYYY}
\acmNumber{X}
\acmDOI{XXXXXXX.XXXXXXX}
\acmISBN{}
\acmConference{Proceedings on Privacy Enhancing Technologies}
\theoremstyle{plain}
\newtheorem{theorem}{Theorem}[section]

\theoremstyle{definition}
\newtheorem{definition}[theorem]{Definition}

\theoremstyle{remark}

\begin{document}

\newcommand{\myparagraph}[1]{
\vspace{8pt}\noindent
\textbf{#1.}
}

\renewcommand*{\figureautorefname}{Fig.}
\renewcommand*{\equationautorefname}{Eq.}
\renewcommand*{\sectionautorefname}{Sec.}

\newcolumntype{Y}{>{\centering\arraybackslash}X}

\title{FedLAP-DP: Federated Learning by Sharing Differentially Private Loss Approximations}

\author{Hui-Po Wang}
\affiliation{%
  \institution{CISPA Helmholtz Center for Information Security}
  \city{}
  \state{}
  \country{Germany}}
\email{hui.wang@cispa.de}

\author{Dingfan Chen}
\affiliation{%
  \institution{CISPA Helmholtz Center for Information Security}
  \city{}
  \state{}
  \country{Germany}}
\email{dingfan.chen@cispa.de}

\author{Raouf Kerkouche}
\affiliation{%
  \institution{CISPA Helmholtz Center for Information Security}
  \city{}
  \state{}
  \country{Germany}}
\email{raouf.kerkouche@cispa.de}

\author{Mario Frtiz}
\affiliation{%
  \institution{CISPA Helmholtz Center for Information Security}
  \city{}
  \state{}
  \country{Germany}}
\email{fritz@cispa.de}

\renewcommand{\shortauthors}{Wang et al.}

\begin{abstract}

 Conventional gradient-sharing approaches for federated learning (FL), such as FedAvg, rely on aggregation of local models and often face performance degradation under differential privacy (DP) mechanisms or data heterogeneity, which can be attributed to the inconsistency between the local and global objectives. To address this issue, we propose FedLAP-DP, a novel privacy-preserving approach for FL. Our formulation involves clients synthesizing a small set of samples that approximate local loss landscapes by simulating the gradients of real images within a local region. Acting as loss surrogates, these synthetic samples are aggregated on the server side to uncover the global loss landscape and enable global optimization. Building upon these insights, we offer a new perspective to enforce record-level differential privacy in FL. A formal privacy analysis demonstrates that FedLAP-DP incurs the same privacy costs as typical gradient-sharing schemes while achieving an improved trade-off between privacy and utility. Extensive experiments validate the superiority of our approach across various datasets with highly skewed distributions in both DP and non-DP settings. Beyond the promising performance, our approach presents a faster convergence speed compared to typical gradient-sharing methods and opens up the possibility of trading communication costs for better performance by sending a larger set of synthetic images. The source is available at \url{https://github.com/hui-po-wang/FedLAP-DP}.
\end{abstract}


\maketitle

\input{articles/intro.tex}
\input{articles/related_work.tex}

\input{articles/background.tex}
\input{articles/method.tex}

\input{articles/analysis}
\input{articles/experiments.tex}
\input{articles/discussion.tex}

\input{articles/conclusion.tex}

\begin{acks}
This work is partially funded by the Helmholtz Association within the project "Trustworthy Federated Data Analytics (TFDA)" (ZT-I-OO1 4), Medizininformatik-Plattform "Privatsphären-schutzende Analytik in der Medizin" (PrivateAIM), grant No. 01ZZ2316G, and Bundesministeriums fur Bildung und Forschung (PriSyn), grant No. 16KISAO29K. The work is also supported by ELSA – European Lighthouse on Secure and Safe AI funded by the European Union under grant agreement No. 101070617. Moreover, The computation resources used in this work are supported by the Helmholtz Association's Initiative and Networking Fund on the HAICORE@FZJ partition. Views and opinions expressed are, however, those of the authors only and do not necessarily reflect those of the European Union or European Commission. Neither the European Union nor the European Commission can be held responsible for them.  Dingfan Chen was partially supported by Qualcomm Innovation Fellowship Europe. 
\end{acks}

\bibliographystyle{ACM-Reference-Format}
\bibliography{reference}

\clearpage
\appendix
\section*{Appendices}

\input{articles/appendix}

\end{document}

%% file: math_commands.tex
\usepackage{amsmath,amsfonts,bm}

\def\eqref#1{equation~\ref{#1}}

\def\1{\bm{1}}

\def\vg{{\bm{g}}}

\def\vw{{\bm{w}}}
\def\vx{{\bm{x}}}

\def\mA{{\bm{A}}}
\def\mB{{\bm{B}}}

\DeclareMathAlphabet{\mathsfit}{\encodingdefault}{\sfdefault}{m}{sl}
\SetMathAlphabet{\mathsfit}{bold}{\encodingdefault}{\sfdefault}{bx}{n}

\def\gD{{\mathcal{D}}}
\def\gE{{\mathcal{E}}}

\def\gM{{\mathcal{M}}}
\def\gN{{\mathcal{N}}}

\def\gR{{\mathcal{R}}}
\def\gS{{\mathcal{S}}}

\def\sR{{\mathbb{R}}}

\newcommand{\E}{\mathbb{E}}
\newcommand{\Ls}{\mathcal{L}}

\DeclareMathOperator*{\argmin}{arg\,min}

%% file: articles/intro.tex
\section{Introduction}
\label{sec:intro}

Federated Learning (FL)~\citep{mcmahan2017communication} is a distributed learning framework that allows participants to train a model collaboratively without sharing their data. 
Predominantly, existing works~\citep{mcmahan2017communication, karimireddy2020scaffold, li2020federated} achieve this by training local models on clients' private datasets for several local epochs and sharing only the averaged gradients with the central server. Despite extensive research over the past few years, these prevalent gradient-based methods still suffer from several challenges~\citep{kairouz2021advances} when training with heterogeneous clients. 
These issues could result in significant degradation in performance and convergence speed, thus making FL difficult to scale up.

Aside from the data heterogeneity that has been notoriously known to interfere with federated optimization~\citep{hsu2019measuring, li2019convergence, karimireddy2020scaffold}, potential privacy breaches can still occur in the FL gradient-sharing scheme. For instance, data reconstruction~\citep{hitaj2017deep, bhowmick2018protection, geiping2020inverting, DLG, IDLG, VFLllgusenix, LLG, li2021label} and membership inference attacks~\citep{nasr2019comprehensive,Property} have been widely studied in FL applications. In response to these emerging privacy risks, differential privacy (DP) has been established as the golden standard for providing theoretical privacy guarantees against potential privacy leakage. The most prevalent algorithm to ensure DP properties, namely DP-SGD~\cite{abadi2016deep}, operates by clipping per-sample gradients and adding Gaussian noise, thus obfuscating the influence of each data record. Despite its guarantee, the noisy gradients introduced by DP-SGD could induce additional heterogeneity, an aspect not thoroughly explored in existing FL literature. This aligns with the observation that DP noise leads to a more significant performance drop in FL problems than in centralized settings and was partially explored by~\citet{yang2023privatefl} in  a personalized FL setting.

The fundamental cause of such degradation is that \emph{the local updates, driven by the distinct objectives of heterogeneous clients, optimize the models towards their local minima instead of the global objective}. This inconsistency drives the (weighted) average models learned by existing approaches to sub-optimal performance or even fails the convergence. Existing efforts have proposed various ideas to address the issue of heterogeneity, including variance-reduction~\citep{karimireddy2020scaffold}, reducing the dissimilarity among clients~\citep{li2020federated, li2021model, wang2020tackling}, or personalized models~\citep{luo2021no, lifedbn, fallah2020personalized}. However, these methods mainly focus on non-DP cases and still heavily rely on biased local updates, which are the root cause of the degradation. Overall, the gap between DP, data heterogeneity, and objective inconsistency has yet to be fully bridged and deserves further investigation.

\begin{figure*}
    \begin{center}
      \includegraphics[trim=0 0.3cm 0 0,clip,width=\linewidth]{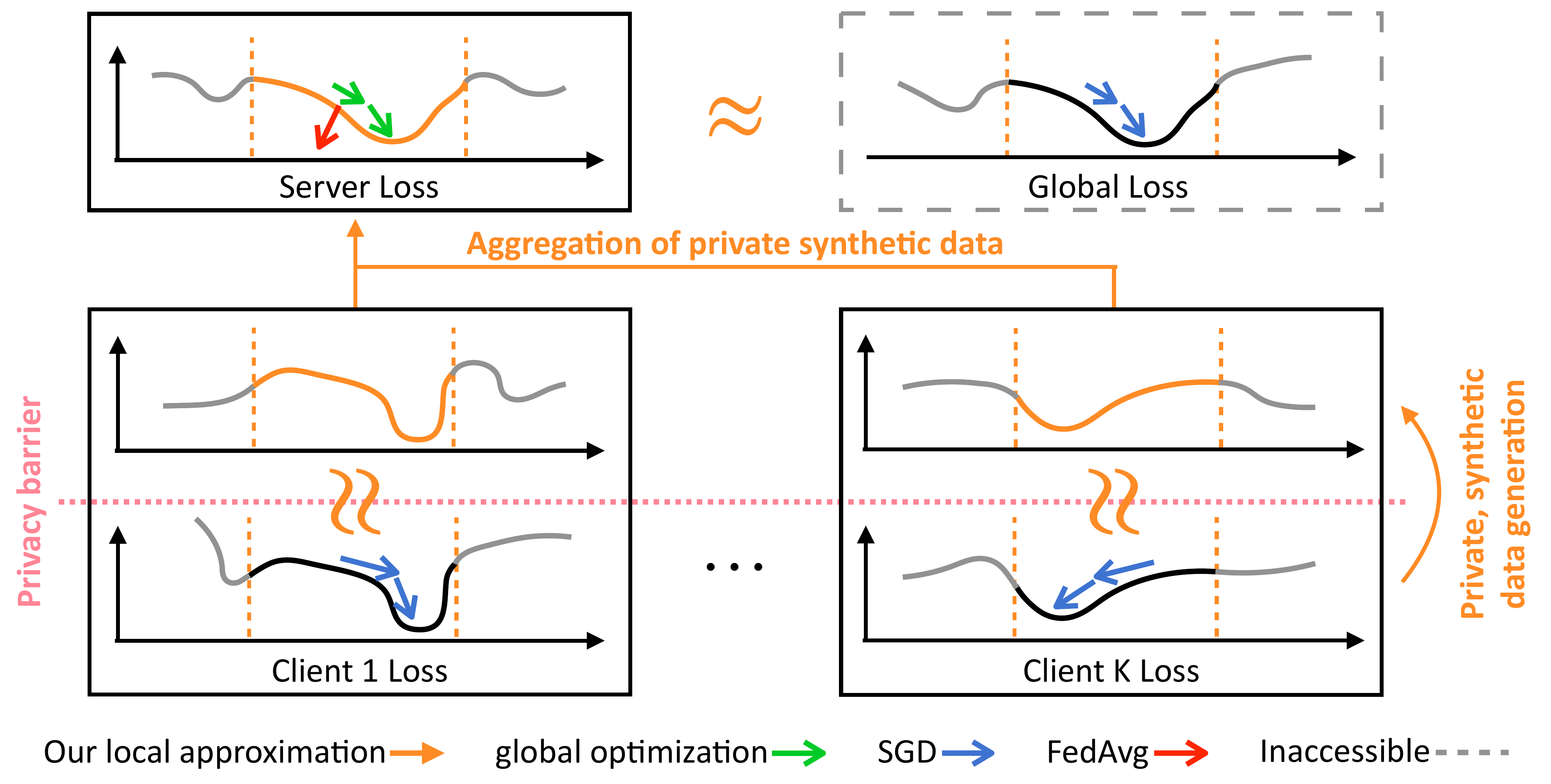}
\caption{An overview of FedLAP-DP. FedLAP-DP addresses the limitations of methods like FedAvg, which aggregate locally optimized gradients (blue) to meet a global objective. Such methods often lead to sub-optimal results (red) due to goal misalignment. This problem further intensifies with increasing client data heterogeneity. FedLAP-DP approximates local neighborhoods with synthetic images on the clients (local approximation, \sectionautorefname~\ref{ssec:local_approx}) and optimizes the model according to the reconstructed loss landscape on the server (global optimization, \sectionautorefname~\ref{ssec:global_opt}). Differential privacy is integrated to introduce privacy barriers (\sectionautorefname~\ref{ssec:method-dp}).}
\label{fig:teaser}
    \end{center}
\end{figure*}

As a pioneering step, this work proposes FedLAP-DP, a novel differentially private framework designed to approximate local loss landscapes and counteract biased federated optimization through the utilization of synthetic samples. As illustrated in \figureautorefname~\ref{fig:teaser}, unlike traditional gradient-sharing schemes~\citep{mcmahan2017communication} that are prone to inherently biased global update directions, our framework transmits synthetic samples encoding the local optimization landscapes. This enables the server to faithfully uncover the global loss landscape, overcoming the biases incurred by conventional gradient-sharing schemes and resulting in substantial improvements in convergence speed (refer to \sectionautorefname~\ref{sec:exp}).  Additionally, we introduce the usage of a trusted region to faithfully reflect the approximation quality, further mitigating bias stemming from potential imperfections in the local approximation within our scheme.

Based on the insights, our approach offers a novel perspective to incorporate record-level differential privacy into federated learning. In particular, we begin with applying DP protection mechanisms, such as DP-SGD, to the gradients produced from real client data. These DP-protected gradients then serve as the learning objective for synthetic dataset optimization. Thanks to the post-process theorem, we are allowed to conduct multiple optimization steps based on the protected real gradients to further improve the approximation quality of the synthetic data without incurring additional privacy costs. Our formal privacy analysis demonstrates that FedLAP-DP consumes the same privacy costs as traditional gradient-sharing baselines. Together with the proposed trusted regions, FedLAP-DP provides reliable utility under privacy-preserving settings, especially when considering low privacy budgets and highly skewed data distributions.

Lastly, we note that our approach has several additional benefits. (1) Our approach is more communication efficient as it enables the execution of multiple optimization rounds on the server side. This is particularly advantageous for large models, where transferring synthetic data is notably less costly than gradients in each round.  (2) While FedLAP-DP exhibits superior performance under the same communication costs as the baselines, our method can increase the convergence speed and further improve performance by synthesizing a larger set of synthetic images. This opens up the possibility of trading additional costs for better performance, which is little explored in existing literature. Overall, we summarize our contributions as follows. \vspace{5pt}

\begin{itemize}[leftmargin=12pt,nosep]
    \item We propose FedLAP-DP that offers a novel perspective of federated optimization by transferring local loss landscapes to uncover the global loss landscape on the server side, avoiding potentially biased local updates and enabling unbiased federated optimization on the server.
    \item We demonstrate how to synthesize samples that approximate local loss landscapes and effective regions on clients, minimizing the effect of imperfect approximation.
    \item Along with a formal privacy analysis, we demonstrate an innovative way to integrate record-level DP into federated learning without incurring additional privacy costs.
    \item Extensive experiments confirm the superiority of our formulation over existing gradient-sharing baselines in terms of performance and convergence speed, particularly when considering tight privacy budgets and highly skewed data distributions.
\end{itemize}

%% file: articles/related_work.tex
\section{Related Work}
\label{sec:related_work}
\myparagraph{Non-IID Data in Federated Learning} The existing gradient-sharing schemes typically begin with training local models on local private data for several local epochs, and the server aggregates those models to update the global model. However, the applications of FL may naturally introduce heterogeneity among clients due to the contrasting behaviors of users. Such heterogeneity inevitably results in the inconsistency between local and global learning objectives, making the (weighted) averages of local models sub-optimal to the global learning task. This issue~\citep{kairouz2021advances} often causes degradation in performance and convergence speed and has attracted a significant amount of attention from the community.

 To address the issue, existing efforts mainly fall into the following categories: variance-reduction techniques~\citep{karimireddy2020scaffold}, constraining the dissimilarity of updates among clients~\citep{li2020federated, li2021model, wang2020tackling}, and adjusting the global model to a personalized version at the inference stage~\citep{luo2021no, lifedbn, fallah2020personalized}. Variance-reduction techniques~\citep{karimireddy2020scaffold} consider an additional variable to explicitly correct the error induced by the heterogeneity. However, the variable consumes additional communication costs and could be costly. For instance, SCAFFOLD~\citep{karimireddy2020scaffold} consumes twice the cost compared to plain FedAvg. On the other hand, several prior works propose various methods to limit the dissimilarity among clients, such as additional regularization terms~\citep{li2020federated, wang2020tackling} and contrastive learning~\citep{li2021model}. Despite the efficacy, these methods still rely on local models that are biased toward the local optimum. Recently, a growing body of literature has investigated personalizing federated models to compensate for the heterogeneity~\citep{luo2021no, lifedbn, fallah2020personalized}. These personalized methods target different applications and may be constrained when the models are deployed for tasks that extend beyond the original scope of the clients' applications, such as in transfer learning scenarios, even though some existing efforts~\citep{galli2023advancing} have been made to make them DP.

 In contrast, our approach takes advantage of loss landscape approximation by synthetic samples, offering a new global perspective to federated optimization. It does not rely on potentially biased locally updated models and outperforms typical gradient-sharing schemes without introducing additional communication costs or personalized models.

\myparagraph{Differential Privacy in Machine Learning} To address privacy concerns, Differential Privacy (DP) has been established as a standard framework for formalizing privacy guarantees, as outlined in Dwork's foundational work \cite{dwork2014algorithmic}. However, implementing DP involves a compromise between model utility and privacy. This is due to the necessity of modifying model training processes with techniques like clipping updates and injecting noise, which can lead to negative effects of the model performance. Current state-of-the-art methods for training models while maintaining high utility under DP primarily use transfer learning. This involves leveraging models pre-trained on extensive public datasets, subsequently fine-tuned with private data \cite{yosinski2014transferable}. These methods have proven effective across various fields where substantial public datasets exist, showing promising results even with limited private data \cite{tobaben2023Efficacy}. Moreover, parameter-efficient fine-tuning techniques like adaptors, including LoRA \cite{hu2022LoRA}, have demonstrated competitiveness in transfer learning within DP contexts%
\cite{yu2022differentially,tobaben2023Efficacy}. Similarly, compression strategies have been investigated to minimize the model's size or its updates. This size reduction consequently lowers the sensitivity and the associated noise introduced by DP \cite{FL_CS_DP,FL_Cons_DP}. However, only a limited number of research studies have explored the application of DP in federated settings characterized by significant heterogeneity among client datasets~\cite{noble2022differentially,liu2021projected, heter_dp_fl}. These approaches, similar to FedAvg, heavily rely on biased local models, which is the root cause of performance degradation.

\myparagraph{Dataset Distillation} 
Our work is largely motivated by recent progress in the field of \textit{dataset distillation}~\citep{wang2018dataset,zhao2021dataset,zhao2023dataset,cazenavette2022dataset}, a process aimed at distilling the necessary knowledge of model training into a small set of synthetic samples. These samples are optimized to serve as surrogates for real data in subsequent model training. The concept of dataset distillation dates back to Wang et al.~\cite{wang2018dataset}, who formulated the dataset distillation as a bi-level optimization problem and introduced a meta-learning approach. In this approach, the inner optimization simulates training a neural network on a distilled synthetic dataset, while the outer optimization updates the synthetic dataset through gradient descent. While this method has shown effectiveness for certain smaller networks, it requires substantial computational resources due to the gradient unrolling operation required in bi-level optimization, making it challenging to be used with larger, more common models.

Recent developments have enhanced this original approach by eliminating the need for complex bi-level optimization. In this context, various directions have been explored, such as gradient matching~\cite{zhao2021dataset}, distribution matching~\cite{zhao2023dataset}, and trajectory matching~\cite{cazenavette2022dataset}. Our approach builds upon gradient matching, which offers more computational efficiency and better compatibility with DP training than trajectory matching. Furthermore, it more effectively exploits discriminative information from a specific global model, making it generally more suitable for FL settings. Unlike distribution matching, which tends to prioritize generalization across various downstream models, gradient matching focuses on specialization for a certain global model, a subtler but more advantageous trade-off for FL training. Specifically, our approach draws inspiration from DSC~\citep{zhao2021dataset} but incorporates several key distinctions to enhance the efficacy and applicability in FL training.  

The proposed FedLAP-DP
(1) focuses on finding local approximation and assembling the global loss landscape to facilitate federated optimization, (2) is class-agnostic and complements record-level differential privacy while prior works often consider class-wise alignment and could cause privacy risks, and (3) is designed for multi-round training with several critical design choices. 
Notably, the most closely related research to ours is PSG~\citep{chenprivate}, which also explored class-agnostic distillation with DP guarantees. However, the focus of PSG is on single-round distillation rather than the standard multi-round federated learning.

%% file: articles/background.tex
\section{Background}
\label{sec:background}

\subsection{Federated Learning}
\label{ssec:fl}
In federated learning, we consider training a model $\mathbf{w}$ that maps the input $\vx$ to the output prediction $y$. We assume $K$ clients participate in the training, and each owns a private dataset $\gD_k$ associated with distribution $p_k$. %
We use the subscript $k$ to represent the indices of clients and the superscript $m$ and $t$ to denote the $m$-th communication round and $t$-th local step, respectively, unless stated otherwise.

Overall, the learning objective is to find the optimal model $\mathbf{w}$ that minimizes the empirical global loss over the population distribution:
\begin{equation}
    \label{eq:global_loss}
    \mathcal{L}(\mathbf{w}) = \E_{(\vx, y) \sim p} [\ell(\mathbf{w}, \vx, y)] = \frac{1}{N}\sum_{j=1}^{N} \ell(\mathbf{w}, \vx_j,y_j)
\end{equation}
where $\ell$ could be arbitrary loss criteria such as cross-entropy and $N$ is the total dataset size. 
However, in federated settings, direct access to the global objective is prohibited as all client data is stored locally. Instead, the optimization is conducted on local surrogate objectives $\Ls_k(\mathbf{w})$:
\begin{align*}
\small
    \mathcal{L}(\mathbf{w}) &= \sum_{k=1}^{K} \frac{N_k}{N} \Ls_k(\mathbf{w}) \, ,\, \Ls_k(\mathbf{w})=\sum_{j=1}^{N_k}\frac{1}{N_k} \ell(\mathbf{w}, \vx_j,y_j),
\end{align*}
where $(\vx_j,y_j)$ are data samples from the client dataset $\gD_k$.

Existing methods, such as FedAvg~\citep{mcmahan2017communication}, simulate stochastic gradient descent on the global objective by performing local gradient updates and periodically aggregating and synchronizing them on the server side.  Specifically, at the $m$-th communication round, the server broadcasts the current global model weights $\mathbf{w}_g^{m,1}$ to each client, who then performs $T$ local iterations with learning rate $\eta$. 

\begin{equation}
    \label{eq:local_update}
    \begin{gathered}
        \mathbf{w}^{m, 1}_k \leftarrow \mathbf{w}^{m, 1}_g, \forall k\in[K] \\
        \mathbf{w}_{k}^{m, t+1} = \mathbf{w}_k^{m, t} - \eta \nabla \mathcal{L}_{k}(\mathbf{w}_{k}^{m, t}), \forall t \in [T]
    \end{gathered}
\end{equation}

The local updates $\Delta \mathbf{w}_k^m$ are then sent back to the server and combined to construct $\widehat{\vg^m}$, which is essentially a linear approximation of the true global update $\vg^m$:

\begin{equation}
\label{eq:approx_grad}
    \begin{gathered}
    \widehat{\vg^m} = \sum_{k=1}^K \frac{N_k}{N} \Delta \mathbf{w}_k^m=\sum_{k=1}^K \frac{N_k}{N} (\mathbf{w}_k^{m,T} - \mathbf{w}_k^{m,1}) \\
    \mathbf{w}_g^{m+1, 1} = \mathbf{w}_g^{m,1} - \eta \widehat{\vg^m}
    \end{gathered}
\end{equation}

In this work, we focus on the conventional FL setting in which each client retains their private data locally, and the local data cannot be directly accessed by the server. Every data sample is deemed private, with neither the server nor the clients using any additional (public) data, which stands in contrast to previous works that require extra data on the server side~\citep{zhao2018federated, li2019fedmd}.
Furthermore, all clients aim towards a singular global objective (\equationautorefname~\ref{eq:global_loss}), which is distinct from personalized approaches wherein evaluations are conducted based on each client's unique objective and their own data distribution~\citep{lifedbn,fallah2020personalized}.

\subsection{Non-IID Challenges}
\label{ssec:non-iid}
The heterogeneity of client data distributions presents several major challenges to FL, such as a significant decrease in the convergence speed (and even divergence) and the final performance when compared to the standard IID training setting~\citep{khaled2019first,li2020federated,karimireddy2020scaffold,li2019convergence}. This can be easily seen from the mismatch between the local objectives that are being solved and the global objective that FL ultimately aims to achieve, i.e., $\Ls_k(\mathbf{w}) \neq \E_{(\vx, y) \sim p} [\ell(\mathbf{w}, \vx, y)] $ if $p_k \neq p$ for some k. Executing multiple local steps on the local objective (Eq.~\ref{eq:local_update}) makes the local update $\Delta \mathbf{w}_k^m$ deviate heavily from the true global gradient $\nabla \Ls (\mathbf{w})$, inevitably resulting in a biased approximation of the global gradient via Eq.~\ref{eq:approx_grad}, i.e., $\widehat{\vg^m} \neq \vg^m$, where $\vg^m$ is derived from the true loss $\Ls(\vw)$ (See \figureautorefname~\ref{fig:teaser} for a demonstration.). 

Despite significant advances achieved by existing works
in alleviating divergence issues, these methods still exhibit a systematic bias generally deviating from optimizing the global objective as they rely on the submitted client updates  $\Delta \mathbf{w}_k^m$, which only reflect a single direction towards the client's
local optimum.

In contrast, our method communicates the synthetic samples $\gS_k$ that encode the local optimization landscapes, i.e., gradient directions within a trust region around the starting point that summarize on possible trajectories $(\mathbf{w}_k^{m,1},\mathbf{w}_k^{m,2},...,\mathbf{w}_k^{m,T+1})$.
This differs significantly from traditional methods that convey only a single direction $\Delta \mathbf{w}_k^m = \mathbf{w}_k^{m,T+1} - \mathbf{w}_k^{m,1}$. This fundamental change provides the central server with a global perspective that faithfully approximates the ground-truth global optimization (See \figureautorefname~\ref{fig:teaser} top row) than existing approaches. %

\subsection{Differential Privacy}
\label{ssec:dp}
Differential Privacy provides theoretical guarantees of privacy protection while allowing for quantitative measurement of utility. We review several definitions used in this work in this section.

\begin{definition}[Differential Privacy~\citep{dwork2014algorithmic}] A randomized mechanism $\gM$ with
range $\gR$ satisfies $(\varepsilon, \delta)$-DP, if for any two adjacent datasets $E$ and $ E'$, i.e., $E'=E \cup \{\vx\}$ for some $\vx$ in the data domain (or vice versa), and for any subset of outputs $O \subseteq \gR$, it holds that
\begin{equation}
    \Pr[\gM(E) \in O] \leq e^\varepsilon \Pr[\gM(E') \in O] + \delta
\end{equation}
\end{definition}

Intuitively, DP guarantees that an adversary, provided with the output of $\mathcal{M}$, %
can only make nearly identical conclusions (within an $\varepsilon$ margin with probability greater than $1 - \delta$) about any specific record, regardless of whether it was included in the input of $\mathcal{M}$ or not~\citep{dwork2014algorithmic}. This suggests that, for any record owner, a privacy breach due to its participation in the dataset is unlikely.

In FL, the notion of \textit{adjacent (neighboring) datasets} used in DP generally refers to pairs of datasets differing by either one user (\textit{user-level} DP) or a single data point of one user (\textit{record-level} DP). Our work focuses on the latter. The definition is as follows.

\begin{definition}[Record-level DP]
A randomized mechanism $\gM$ with range $\gR$ is said to satisfy $(\varepsilon, \delta)$ record-level DP in the context of federated learning if, for any pair of adjacent datasets $E$ and $E'$ differing by a single data point on a client (i.e., $E' = E \cup \{ \vx \}$), the mechanism satisfies $(\varepsilon, \delta)$-DP. Here, $E = \bigcup E_i$ denotes the union of client datasets.
\end{definition}

While there are established methods providing record-level DP for training federated models~\citep{truex2019hybrid,peterson2019private,kerkouche2021privacy}, these primarily operate on the transmitted single client gradients. In contrast, our novel formulation allows efficient communication of comprehensive information, thereby circumventing biased optimization and displaying improved training stability and utility.

We use the Gaussian mechanism to upper bound privacy leakage when transmitting information from clients to the server.  
\begin{definition}[Gaussian Mechanism~\citep{dwork2014algorithmic}]
Let $f: \sR^n \rightarrow \mathbb{R}^d$ be an arbitrary function %
with sensitivity being the maximum Euclidean distance between the outputs over all adjacent datasets $E$ and $E'$:
\begin{equation}
  \Delta_2 f= \max_{E,E'} \Vert f(E)-f(E')\Vert_2  
  \label{def:sensitivity}
\end{equation}
The Gaussian Mechanism  $\mathcal{M}_\sigma$, parameterized by $\sigma$, adds noise into the output, i.e.,
\begin{equation}
  \mathcal{M_\sigma}(\vx) = f(\vx) + \mathcal{N}(0,\sigma^2 \mathbb{I}).
\end{equation}
$\gM_\sigma$ is $(\varepsilon,\delta)$-DP for $\sigma\geq \sqrt{2\ln{(1.25/\delta)}}\Delta_2 f/\varepsilon$. %
\label{def:gaussian_mechanism}
\end{definition}

\begin{theorem}
[Post-processing~\citep{dwork2014algorithmic}] \label{theorem:post-processing} If $\mathcal{M}$  satisfies $(\varepsilon,\delta)$-DP, $G\circ \mathcal{M}$ will satisfy $(\varepsilon,\delta)$-DP for any data-independent function $G$.
\end{theorem}

Moreover, we use Theorem~\ref{theorem:post-processing} to guarantee that the privacy leakage is bounded upon obtaining gradients from real private data in our framework. This forms the basis for the overall privacy guarantee of our framework and enables us to enhance the approximation quality without introducing additional privacy costs.

%% file: articles/method.tex
\begin{algorithm}[tb]
    \caption{FedLAP: Local Approximation}
    \label{alg:nondp-client}
    \newcommand{\Linecomment}[1]{\STATE {\textcolor{blue}{/* #1 */}}}
    \renewcommand{\algorithmicendfunction}{\textbf{Return: }}
    \renewcommand{\algorithmiccomment}[1]{\textcolor{blue}{/* #1 */}}
    \begin{algorithmic}
        \FUNCTION{ClientExecute($k$, $r$, $\mathbf{w}^{m, 1}_g$) :}
        \STATE{\textbf{Initialize} $\gS_k$: $\{\hat{\vx}_k^{m}\}$ from Gaussian noise or $\{\hat{\vx}_k^{m-1}\}$, $\{\hat{y}_k\}$ to be a balanced set }
        \FOR{$i=1, \dots, R_i$ } \Linecomment{Resample training trajectories} 
            \STATE {Reset $t \gets 1$, model $\mathbf{w}^{m, 1}_k \leftarrow \mathbf{w}^{m, 1}_g$, and $\gS_k^{i,0} \gets \gS_k^{i-1}$} 
            \WHILE{$|| \mathbf{w}^{m, t}_k - \mathbf{w}^{m, 1}_k|| < r$}
            \STATE{Sample real data batches $\{(\mathbf{x}_k, y_k)\}$ from $\gD_k$}
            \STATE{ Compute $g^\gD =\nabla\Ls(\mathbf{w}_k^{m,t},\{(\mathbf{x}_k, y_k)\})$}
            \FOR{$j=1, \dots, R_b$}
            \Linecomment{Update synthetic set $\gS_k$ given the real gradient}
                \STATE{
                $\gS_k^{i,j+1}$ = $\gS_k^{i,j} - \tau\nabla_{\gS_k} \Ls_{dis}\Big(g^\gD, \nabla\Ls(\mathbf{w}_k^{m,t},\gS_k^{i,j})\Big)$
    }
            \ENDFOR
            \FOR{$l=1, \dots, R_l$}
            \Linecomment{Update local models from $\mathbf{w}_k^{m,t}$ to $\mathbf{w}_k^{m,t+l}$}
                \STATE{
                $\mathbf{w}^{m ,t+1}_k = \mathbf{w}^{m ,t}_k - \eta \nabla \mathcal{L}(\mathbf{w}^{m ,t}_k, \gS_k)$}
                \STATE $t \gets t+1$
            \ENDFOR
            \ENDWHILE
        \ENDFOR
        \STATE Measure $r_k$ on $\gD_k$ (See \figureautorefname~\ref{fig:epsilon})
        \ENDFUNCTION Synthetic set $\gS_k^{R_i}$, calibrated radius $r_k$
    \end{algorithmic}
\end{algorithm}

\section{FedLAP-DP}
\label{sec:method}

\subsection{Overview}
\label{ssec:overview}

In this section, we start by introducing FedLAP, the non-DP variant of FedLAP-DP, to demonstrate the concept of loss approximation and global optimization. Next, we discuss the effective integration of record-level DP into Federated Learning (FL). This will be followed by a detailed privacy analysis presented in \sectionautorefname~\ref{sec:priv_analysis}.

Contrary to conventional methods that generally transmit local update directions to estimate the global objective (\equationautorefname~\ref{eq:approx_grad}), FedLAP uniquely simulates global optimization by sending a compact set of synthetic samples. These samples effectively represent the local loss landscapes, as illustrated in \figureautorefname~\ref{fig:teaser}.

Let $p_k$ and the $p_{\gS_k}$ be the distribution of the real client dataset $\gD_k$ and the corresponding synthetic dataset $\gS_k$, respectively. We formalize our objective and recover the global objective as follows:
 \begin{equation}
 \label{eq:recover_global}
 \begin{gathered}
 \E_{(\vx,y)\sim p_k} [\ell (\mathbf{w}, \vx, y)] \simeq \E_{(\hat{\vx},\hat{y})\sim p_{\gS_k}}[\ell(\mathbf{w},\hat{\vx}, \hat{y}) ] \\ 
 \Ls (\mathbf{w}) = \sum_{k=1}^K \frac{N_k}{N}\Ls_k(\mathbf{w}) \simeq \sum^K_{k=1} \frac{N_k}{N}{\widehat{\Ls}}_k(\mathbf{w})
 \end{gathered}
 \end{equation}
Thus, performing global updates is then equivalent to conducting vanilla gradient descent on the recovered global objective, i.e., by training on the synthetic set of samples. 

We demonstrate our framework in \figureautorefname~\ref{fig:teaser}. In every communication round, synthetic samples are optimized to approximate the client's local loss landscapes (Sec~\ref{ssec:local_approx}) and then transmitted to the server. The server then performs global updates on the synthetic samples to simulate global optimization (\sectionautorefname~\ref{ssec:global_opt}). Finally, \sectionautorefname~\ref{ssec:method-dp} explains the integration of \textit{record-level} differential privacy into Federated Learning, resulting in the creation of FedLAP-DP. 

The overall algorithm is depicted in Algorithm~\ref{alg:nondp-client} and~\ref{alg:nondp-server} for non-DP settings and Algorithm~\ref{alg:dp} for DP settings, where the indices $i$ being the number of training trajectories observed by the synthetic images and $j$ being the number of updates on a sampled real batch.  For conciseness, we omit the indices in the following where their absence does not affect clarity or understanding.

\begin{figure}[tb]
    \centering
    \includegraphics[width=\columnwidth]{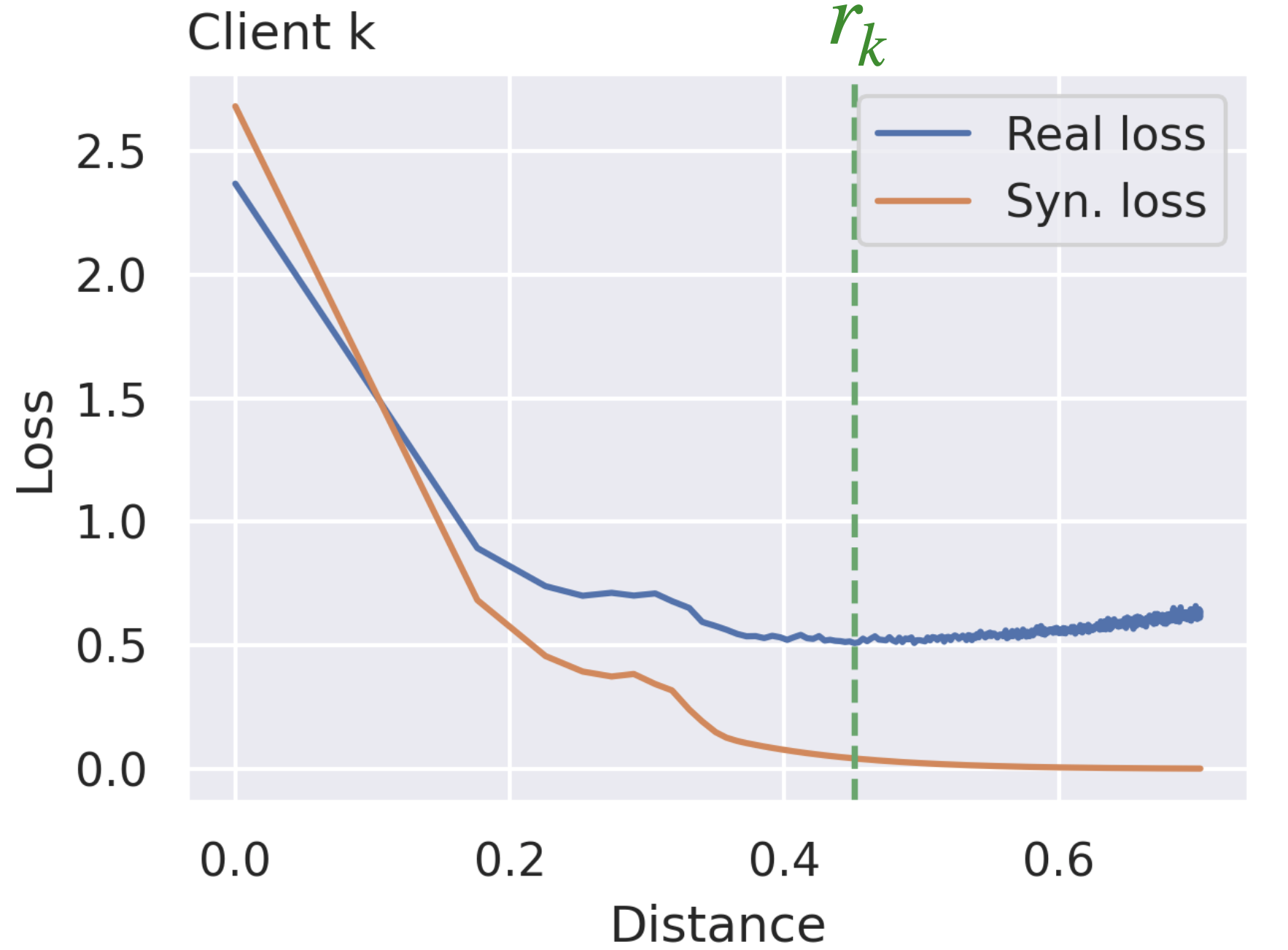}
    \caption{$r_k$ selection. The loss on private real and synthetic data decreases initially but deviates later. $r_k$ is defined as the turning points with the smallest real loss.}
    \label{fig:epsilon}
\end{figure}

\subsection{Local Approximation}
\label{ssec:local_approx}
The goal of this step is to construct a set of synthetic samples $\gS_k$ that accurately captures necessary local information for subsequent global updates. A natural approach would be to enforce similarity between the gradients obtained from the real client data and those obtained from the synthetic set:
 \begin{equation*}
\nabla_\mathbf{w}\E_{(\vx,y)\sim p_k} [\ell (\mathbf{w}, \vx, y)] \simeq \nabla_\mathbf{w}\E_{(\hat{\vx},\hat{y})\sim p_{\gS_k}}[\ell(\mathbf{w},\hat{\vx}, \hat{y}) ]
 \end{equation*}
 We achieve this by minimizing the distance between the gradients:
 \begin{equation}
     \argmin_{\gS_k}\quad \Ls_{\mathrm{dis}}\Big(\nabla\Ls(\mathbf{w},\gD_k), \nabla\Ls(\mathbf{w},\gS_k)\Big)
     \label{eq:local_approx_raw}
 \end{equation}
where $\nabla \Ls (\mathbf{w},\gD_k)
)$ denotes the stochastic gradient of network parameters on the client dataset $\gD_k$, and $\nabla \Ls(\mathbf{w},\gS_k)$ the gradient on the synthetic set for brevity. $\Ls_{\mathrm{dis}}$ can be arbitrary metric that measures the similarity. 

\myparagraph{Distance Metric} 
In this study, we utilize a layer-wise cosine distance as described by \citet{zhao2021dataset}. Furthermore, we have empirically determined that incorporating a mean square error term, which captures directional information and controls the differences in magnitudes, enhances both stability and approximation quality. We provide a more comprehensive explanation and analysis in \sectionautorefname~\ref{ssec:appendix-matching-criteria}. The combination of the distance metric and the mean square error term is formalized as follows.

\begin{equation}
\label{eq:local-loss}
    \begin{gathered}
        \Ls_{\mathrm{dis}} \Big(\nabla_\mathbf{w} \Ls(\mathbf{w},\gD_k), \nabla_\mathbf{w}  \Ls (\mathbf{w}, \gS_k)\Big) \\ 
        = \sum_{l=1}^L d\Big(\nabla_{\mathbf{w}^{(l)}} \Ls(\mathbf{w}^{(l)},\gD_k
        ), \nabla_{\mathbf{w}^{(l)}} \Ls (\mathbf{w}^{(l)},\gS_k)\Big) \\ 
        + \lambda\Vert \nabla_{\mathbf{w}^{(l)}} \Ls(\mathbf{w}^{(l)},\gD_k
        ) - \nabla_{\mathbf{w}^{(l)}} \Ls (\mathbf{w}^{(l)},\gS_k) \Vert_2^2
    \end{gathered},
\end{equation}

where $\lambda$ is a hyper-parameter controlling the strength of regularization, and $d$ denotes the cosine distance between the gradients at each layer:

\begin{equation}
    \label{eq:cosine_similarity}
    d(\mA,\mB) = \sum_{i=1}^{out} \left( 1-\frac{\mA_{i\cdot}\cdot \mB_{i\cdot}}{\Vert \mA_{i\cdot} \Vert \Vert \mB_{i\cdot}\Vert}\right)
\end{equation}

$\mA_{i\cdot}$ and $\mB_{i\cdot}$ represent the flattened gradient vectors corresponding to each output node $i$. 
In fully-connected layers, the parameters $\vw^{(l)}$ inherently form a 2D tensor, whereas the parameters of convolutional layers are represented as 4D tensors. To compute the loss, we flatten the last three dimensions, converting them into 2D tensors as well.
denote the number of output and input channels, kernel height, and width, respectively.

 \myparagraph{Effective Approximation Regions} While solving \equationautorefname~\ref{eq:local_approx_raw} for every possible $\mathbf{w}$ would lead to perfect recovery of the ground-truth global optimization in principle. However, it is practically infeasible due to the large space of (infinitely many) possible values of $\mathbf{w}$. Additionally, as $|\gS_k|$ is set to be much smaller than $N_k$ (for the sake of communication efficiency), an exact solution may not exist, resulting in approximation error for some $\mathbf{w}$. To address this, we explicitly constrain the problem space to be the most achievable region for further global updates. Specifically,
we consider $\mathbf{w}_k$ that is sufficiently close to the initial point of the local update and is located on the update trajectories (\equationautorefname~\ref{eq:local_approx_cst}). Formally,
\begin{align}
\small
    \label{eq:local_approx}
    \argmin_{\gS_k} \, & \sum_{t=1}^T    \Ls_{\mathrm{dis}}\Big(\nabla\Ls(\mathbf{w}_k^{m,t},\gD_k), \nabla\Ls(\mathbf{w}_k^{m,t},\gS_k)\Big) \\
    \label{eq:local_approx_cst}
    \textrm{s.t.} \quad & \| \mathbf{w}^{m ,t}_k - \mathbf{w}^{m ,1}_k \| < r, \\
    &\mathbf{w}^{m ,t+1}_k = \mathbf{w}^{m ,t}_k - \eta \nabla \mathcal{L}(\mathbf{w}^{m ,t}_k, \gS_k)
\end{align}
where $r$ represents a radius suggested by the server, defining the coverage of update trajectories, and $\eta$ denotes the model update learning rate shared among the server and clients.

 In the $m$-th communication round, the clients first synchronize the local model $\mathbf{w}^{m, 1}_k$ with the global model $\mathbf{w}^{m, 1}_g$ and initialize the synthetic features $\{\hat{\mathbf{x}}_k^m\}$ either from  Gaussian noise or to be the ones obtained from the previous round  $\{\hat{\mathbf{x}}_k^{m-1}\}$. Synthetic labels $\{\hat{y}_k\}$ are initialized to be a fixed, balanced set and are not optimized during the training process. The number of synthetic samples $|\gS_k|$ is kept equal for all clients in our experiments, though it can be adjusted for each client depending on factors such as local dataset size and bandwidth in practice.

 To simulate the local training trajectories, the clients alternate between updating synthetic features using \equationautorefname~\ref{eq:local_approx} and updating the local model using \equationautorefname~\ref{eq:local_approx_cst}. This process continues until the current local model weight $\mathbf{w}_k^{m,t}$ exceeds a predefined region $r$ determined by the Euclidean distance on the flattened weight vectors, meaning it is no longer close to the initial point. On the other hand, the server optimization should take into consideration the approximation quality of $\gS_k$. Thus, as illustrated in \figureautorefname~\ref{fig:epsilon}, each client will suggest a radius $r_k$ indicating the distance that $\gS_k$ can approximate best within the radius $r$. For the DP training setting, we make the choice of $r_k$ data-independent by setting it to be a constant (the same as $r$ in our experiments).

\myparagraph{Details of Algorithm} 
The process is outlined in Algorithm~\ref{alg:nondp-client}. In detail, the local model $w^{m,t}$ is reset to the initial global model $w^{m, 1}$ for $R_i$ iterations. This step explores various training paths, enhancing generalizability. Synthesis occurs when the current model $w^{m,t}$ is within a certain distance, defined as the server-suggested radius $r$, from the initial point $w^{m, 1}$. For each real batch, we update synthetic images $R_b$ times and the local model $R_l$ times, moving from local step $t$ to $t+R_l$. Finally, the client assesses the effective approximation regions and communicates the radius $r_k$. Building on previous studies~\citep{zhao2021dataset}, we note that the constraints are not enforced through classical means of constrained optimization, like KKT conditions. Rather, they are applied via the termination of loops, a method necessitated by the infeasibility of exact solutions.

\begin{algorithm}[tb]
\caption{FedLAP: Global Optimization}
 \label{alg:nondp-server}
\newcommand{\Linecomment}[1]{\STATE {\textcolor{blue}{/* #1 */}}}
\renewcommand{\algorithmicendfunction}{\textbf{Return: }}
\renewcommand{\algorithmiccomment}[1]{\textcolor{blue}{/* #1 */}}
\begin{algorithmic}
\FUNCTION{ServerExecute:}
    \STATE {\textbf{Initialize} global weight $\mathbf{w}^{1, 1}_g$, radius $r$}
    \Linecomment{Local approximation}
    \FOR{$m = 1, \dots, M$} 
        \FOR{$k = 1, \dots, K$}
            \STATE $\gS_k, r_k \gets $ ClientExecute($k$, $r$, $\mathbf{w}^{m, 1}_g$)
        \ENDFOR
        \Linecomment{Global optimization}
        \STATE {$r_g \gets \min \{r_k\}_{k=1}^K$}
        \STATE {$t \gets 1$}
    \WHILE{$\Vert\mathbf{w}^{m, 1}_g -\mathbf{w}^{m, t}_g\Vert < r_g$}
        \STATE{
        \scalebox{0.9}[0.9]{
        $\mathbf{w}_g^{m,t+1} = \mathbf{w}_g^{m,t} - \sum^K_{k=1} \eta\frac{N_k }{N}
    \nabla \Ls(\mathbf{w}_g^{m,t},\gS_k)$}
    }
        \STATE $t \gets t+1$
        \ENDWHILE
        \STATE $\mathbf{w}^{m+1, 1}_g \gets \mathbf{w}^{m, t}_g$
    \ENDFOR
    \ENDFUNCTION global model weight $\mathbf{w}_g^{M+1,1}$
\end{algorithmic}
\end{algorithm}

\subsection{Global Optimization}
\label{ssec:global_opt}
Once the server received the synthetic set $\gS_k$ and the calibrated radius $r_k$, global updates can be performed by conducting gradient descent directly on the synthetic set of samples. The global objective can be recovered by $\widehat{\Ls}_k(\mathbf{w})$ according to \equationautorefname~\ref{eq:recover_global} (i.e., training on the synthetic samples), while the scaling factor $\frac{N_k}{N}$ can be treated as the scaling factor of the learning rate when computing the gradients on samples from each synthetic set $\gS_k$, namely:
\begin{equation}
\label{eq:global_opt_cst}
    \begin{gathered}
        \mathbf{w}_g^{m,t+1} = \mathbf{w}_g^{m,t} - \sum^K_{k=1} \eta \cdot \frac{N_k }{N}
        \nabla_{\mathbf{w}} \Ls(\mathbf{w}_g^{m,t},\gS_k) \\ 
        \textrm{s.t.} \quad  \Vert \mathbf{w}^{m,t}_g - \mathbf{w}^{m,1}_g \Vert \leq \min\, \{r_k\}_{k=1}^K
    \end{gathered}
\end{equation}
The constraint in \equationautorefname~\ref{eq:global_opt_cst} enforces that the global update respects the vicinity suggested by the clients, meaning updates are only made within regions where the approximation is sufficiently accurate.

\begin{algorithm}[tb]
    \caption{FedLAP-DP}
    \label{alg:dp}
    \newcommand{\Linecomment}[1]{\STATE {\textcolor{blue}{/* #1 */}}}
    \renewcommand{\algorithmicendfunction}{\textbf{Return: }}
    \renewcommand{\algorithmiccomment}[1]{\textcolor{blue}{/* #1 */}}
    \begin{algorithmic}
        \FUNCTION{ServerExecute:}
        \STATE {\textbf{Initialize} global weight $\mathbf{w}^{1, 1}_g$, Fix the radius $r$}
        \Linecomment{Local approximation}
        \FOR{$m = 1, \dots, M$} 
            \FOR{$k = 1, \dots, K$}
                \STATE $\gS_k \gets $ ClientsExecute($k$, $r$, $\mathbf{w}^{m, 1}_g$)
            \ENDFOR
            \Linecomment{Global optimization}
            \STATE {$t \gets 1$}
        \WHILE{$\Vert\mathbf{w}^{m, 1}_g -\mathbf{w}^{m, t}_g\Vert < r$}
            \STATE{
            $\mathbf{w}_g^{m,t+1} = \mathbf{w}_g^{m,t} - \sum^K_{k=1} \eta\frac{1}{K}
        \nabla_{\mathbf{w}} \Ls(\mathbf{w}_g^{m,t},\gS_k)$
        }
            \STATE $t \gets t+1$
            \ENDWHILE
            \STATE $\mathbf{w}^{m+1, 1}_g \gets \mathbf{w}^{m, t}_g$
        \ENDFOR
        \ENDFUNCTION global model weight $\mathbf{w}_g^{M+1,1}$
    \\\hrulefill
        \FUNCTION{ClientExecute($k$, $r$, $\mathbf{w}^{m, 1}_g$) :}
        \STATE{\textbf{Initialize} $\gS_k$: $\{\hat{\vx}_k^{m}\}$ from Gaussian noise or $\{\hat{\vx}_k^{m-1}\}$, $\{\hat{y}_k\}$ to be a balanced set }
        \FOR{$i=1, \dots, R_i$ } \Linecomment{Resample training trajectories} 
            \STATE {Reset $t \gets 1$, model $\mathbf{w}^{m, 1}_k \leftarrow \mathbf{w}^{m, 1}_g$, and $\gS_k^{i,0} \gets \gS_k^{i-1}$} 
            \WHILE{$|| \mathbf{w}^{m, t}_k - \mathbf{w}^{m, 1}_k|| < r$}
            \STATE{Sample random batch $\{(\vx_k^i, y_k^i)\}_{i=1}^\mathbb{B}$ from $\gD_k$}
            \FOR{each $(\vx_k^i, y_k^i)$}
            \Linecomment{Compute per-example gradients on client data}
            \STATE{$g^\gD (\vx_k^i) =\nabla_\mathbf{w} \ell (\mathbf{w}_k^{m,t}, \vx_k^i, y_k^i)$} 
            \Linecomment{Clip gradients with bound $\mathbb{C}$}\STATE{$\widetilde{g^\gD}(\vx_k^i) = g^\gD(\vx_k^i)\cdot \min(1,\mathbb{C}/\Vert g^\gD(\vx_k^i) \Vert_2)$ }
            \ENDFOR
            \Linecomment{Add noise to average gradient by Gaussian mechanism}
            \STATE{$\nabla \widetilde{\Ls}(\mathbf{w}_k^{m,t}, \gD_k) = \frac{1}{\mathbb{B}}\sum_{i=1}^\mathbb{B}(\widetilde{g^\gD}(\vx_k^i) + \gN(0,\sigma^2 \mathbb{C}^2 I))$}
            \FOR{$j=1, \dots, R_b$}
            \Linecomment{Update synthetic set $\gS_k$}
                \STATE{
                $\gS_k^{i,j+1}$ = $\gS_k^{i,j} - \tau\nabla_{\gS_k} \Ls_{dis}\Big(\nabla\widetilde{\Ls}(\mathbf{w}_k^{m,t}, \gD_k) , \nabla\Ls(\mathbf{w}_k^{m,t},\gS_k)\Big)$
    }
            \ENDFOR
            \FOR{$l=1, \dots, R_l$}
            \Linecomment{Update local model parameter $\mathbf{w}_k$}
                \STATE{%
                $\mathbf{w}^{m ,t+1}_k = \mathbf{w}^{m ,t}_k - \eta \nabla \mathcal{L}(\mathbf{w}^{m ,t}_k, \gS_k)$}
                \STATE $t \gets t+1$
            \ENDFOR
            \ENDWHILE
        \ENDFOR
        \ENDFUNCTION Synthetic set $\gS_k^{R_i}$
    \end{algorithmic}
\end{algorithm}

\subsection{Record-level DP}
\label{ssec:method-dp}

While federated systems offer a basic level of privacy protection, recent works identify various vulnerabilities under the existing framework, such as membership inference~\citep{nasr2019comprehensive,Property}. Though \citet{dong2022privacy} uncovers that distilled datasets may naturally introduce privacy protection, we further address possible privacy concerns that might arise during the transfer of synthetic data in our proposed method. Specifically, we rigorously limit privacy leakage by integrating record-level DP, a privacy notion widely used in FL applications. This is especially important in cross-silo scenarios, such as collaborations between hospitals, where each institution acts as a client, aiming to train a predictive model and leveraging patient data with varying distributions across different hospitals while ensuring strict privacy protection for patients.

\myparagraph{Threat Model}
In a federated system, there can be one or multiple colluding adversaries who have access to update vectors from any party during each communication round. These adversaries may have unlimited computation power but remain passive or "honest-but-curious," meaning they follow the learning protocol faithfully without modifying any update vectors~\citep{truex2019hybrid,peterson2019private,kerkouche2021privacy}. These adversaries can represent any party involved, such as a malicious client or server, aiming to extract information from other parties. The central server possesses knowledge of label classes for each client's data, while clients may or may not know the label classes of other clients' data. While we typically do not intentionally hide label class information among clients, our approach is flexible and can handle scenarios where clients want to keep their label class information confidential from others.
\vspace{10pt}

We integrate record-level DP into FedLAP to provide 
theoretical privacy guarantees, which yields FedLAP-DP. Given a desired privacy budget $(\varepsilon, \delta)$, we clip the gradients derived from real data with the Gaussian mechanism, denoted by $\nabla\widetilde{\Ls}(\mathbf{w}_k^{m,t},\gD_k)$. The DP-guaranteed local approximation can be realized by replacing the learning target of \equationautorefname~\ref{eq:local_approx} with the gradients processed by DP while leaving other constraints the same. Formally, we have
\begin{equation}
    \label{eq:dp_local_loss}
    \begin{split}
    \argmin_{\gS_k} \, & \sum_{t=1}^T    \Ls_{\mathrm{dis}}\Big(\nabla\widetilde{\Ls}(\mathbf{w}_k^{m,t},\gD_k), \nabla\Ls(\mathbf{w}_k^{m,t},\gS_k)\Big) \\
    \textrm{s.t.} \quad & \| \mathbf{w}^{m ,t}_k - \mathbf{w}^{m ,1}_k \| < r, \\
    & \mathbf{w}^{m ,t+1}_k = \mathbf{w}^{m ,t}_k - \eta \nabla \mathcal{L}(\mathbf{w}^{m ,t}_k, \gS_k)
    \end{split}
\end{equation}

During the server optimization, we do not use a data-dependent strategy to decide $r$. Instead, we use the same radius as adopted for local approximation (c.f. \equationautorefname~\ref{eq:global_opt_cst}).

\begin{equation}
\label{eq:dp-global_opt_cst}
    \begin{gathered}
        \mathbf{w}_g^{m,t+1} = \mathbf{w}_g^{m,t} - \sum^K_{k=1} \eta \cdot \frac{N_k }{N}
        \nabla_{\mathbf{w}} \Ls(\mathbf{w}_g^{m,t},\gS_k) \\ 
        \textrm{s.t.} \quad  \Vert \mathbf{w}^{m,t}_g - \mathbf{w}^{m,1}_g \Vert \leq r
    \end{gathered}
\end{equation}

We describe the full algorithm of FedLAP-DP in Algorithm~\ref{alg:dp}
and present the privacy analysis of FedLAP-DP in the \sectionautorefname~\ref{sec:priv_analysis}. Our analysis suggests that with equivalent access to private data, FedLAP incurs the same privacy costs as gradient-sharing approaches. Our method further demonstrates a better privacy-utility trade-off in \sectionautorefname~\ref{ssec:exp_dp}, confirming its robustness under DP noise.

%% file: articles/analysis.tex
\begin{table*}[tb]
\small
\aboverulesep=0.1ex
\belowrulesep=0.1ex
\begin{tabularx}{\linewidth}{@{}XY|YYYYYY@{}}
\toprule
 & DSC$^\dag$ & FedSGD ($1\times$) & FedAvg ($1\times$) & FedProx ($1\times$) & SCAFFOLD ($2\times$) & FedDM ($0.96\times$) & Ours ($0.96\times$) \\ \midrule
MNIST & 98.90±0.20 & 87.07±0.65 & 96.55±0.21 & 96.26±0.04 & \underline{97.56±0.06} & 96.66±0.18 & \textbf{98.08±0.02} \\
Fa.MNIST & 83.60±0.40 & 75.10±0.16 & 79.67±0.56 & 79.37±0.29 & 82.17±0.37 & \underline{83.10±0.16} & \textbf{87.37±0.09} \\
CIFAR-10 & 53.90±0.50 & 60.91±0.19 & \textbf{75.20±0.12} & 63.84±0.45 & 56.27±1.19 & 70.51±0.45 & \underline{71.91±0.20} \\ \bottomrule
\end{tabularx}
\caption{Performance comparison on benchmark datasets. The relative communication cost of each method (w.r.t. the model size) is shown in brackets. DSC$^\dag$ is ported from the original paper and conducted in a one-shot centralized setting.} \vspace{-0.5cm}
\label{tab:perf_comparison}
\end{table*}

\begin{figure*}[tb]
\centering
\includegraphics[width=\linewidth]{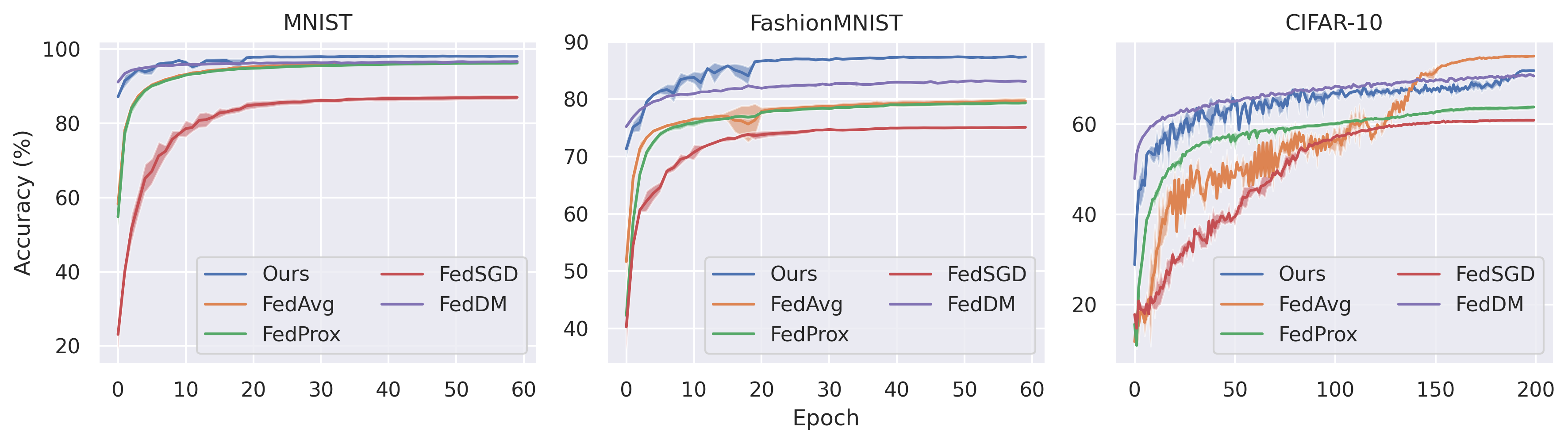}
\caption{Accuracy over communication rounds with extremely non-IID data.}
\label{fig:non-iid-results}
\end{figure*}

\section{Privacy Analysis}

\label{sec:priv_analysis}

\subsection{Definitions}

\begin{definition}[R\'enyi divergence]

Let $P$ and $Q$ be two distributions defined over the same probability space $\mathcal{X}$. Let their respective densities be denoted as $p$ and $q$. The R\'enyi divergence, of a finite order $\alpha \neq 1$, between the distributions $P$ and $Q$ is defined as follows:
\begin{equation}
    D_\alpha\left( P \parallel Q \right) \overset{\Delta}{=} \frac{1}{\alpha -1} \ln \int_{\mathcal{X}} q(x) \left( \frac{p(x)}{q(x)} \right)^\alpha \mathop{dx}. \nonumber
\end{equation}

R\'enyi divergence at orders $\alpha=1,\infty$ are defined by continuity.

\end{definition}

\vspace{10pt}

\begin{definition}[R\'enyi differential privacy (RDP)~\cite{mironov2017renyi}]
\label{def:rdp}
A randomized mechanism $\gM: \gE \rightarrow \gR$ satisfies $(\alpha, \rho)$-R\'enyi differential privacy (RDP) if for any two adjacent inputs $E$, $E' \in \gE$ it holds that

\begin{equation}
  D_\alpha\left( \gM(E) \parallel \gM(E')  \right) \leq \rho   \nonumber
\end{equation}
    
\end{definition}

In this work, we call two datasets $E, E'$ to be adjacent if $E'= E \cup \{ \vx \}$ (or vice versa).

\vspace{10pt}
\begin{definition}[Sampled Gaussian Mechanism (SGM)~\cite{abadi2016deep,mironov2019r}]
\label{def:sgm}
    Let $f$ be an arbitrary function mapping subsets of $\mathcal{E}$ to $\mathbb{R}^d$. We define the Sampled Gaussian mechanism (SGM) parametrized with the sampling rate $0 < q \leq 1$ and the noise $\sigma>0$ as
    \begin{equation}
     \mathrm{SG}_{q,\sigma} \overset{\Delta}{=}  f \left(\{ \vx : \vx \in E \text{ is sampled with probability } q \} \right) + \gN(0,\sigma^2\mathbb{I}_d),\nonumber      
    \end{equation}
where each element of $E$ is independently and randomly sampled  with probability $q$ without replacement. 

The sampled Gaussian mechanism consists of adding i.i.d Gaussian noise with zero mean and variance $\sigma^2$ to each coordinate of the true output of $f$.
In fact, the sampled Gaussian mechanism draws random vector values from a
multivariate isotropic Gaussian distribution denoted by $\gN(0,\sigma^2\mathbb{I}_d)$, where $d$ is omitted if it is unambiguous in the given context.

\end{definition}

\subsection{Analysis}
The privacy analysis of FedLAP-DP and other DP baselines follows the analysis framework used for gradient-based record level-DP methods in FL~\cite{truex2019hybrid,peterson2019private,kerkouche2021privacy}. In this framework, each individual local update is performed as a single SGM (Definition~\ref{def:sgm}) that involves clipping the per-example gradients on a local batch and subsequently adding Gaussian noise to the averaged batch gradient (Algorithm~\ref{alg:dp}). The privacy cost accumulated over multiple local updates and global rounds is then quantified utilizing the revisited moment accountant~\cite{mironov2019r}, which presents an adapted version of the moments accountant introduced in \citet{abadi2016deep} by adapting to the notion of RDP (Definition~\ref{def:rdp}). Finally, to obtain interpretable results and enable transparent comparisons to established approaches, we convert the privacy cost from $(\alpha,\rho)$-RDP to $(\varepsilon, \delta)$-DP by employing Theorem~\ref{th:conv} provided by \cite{balle2020hypothesis}.

\begin{theorem} (\citet{mironov2019r}).
\label{theorem:one-dim} Let $\mathrm{SG}_{q,\sigma}$ be the Sampled Gaussian mechanism for some function $f$ with $\Delta_2 f \leq 1$ for any adjacent $E, E' \in \mathcal{E}$. Then $\mathrm{SG}_{q,\sigma}$ satisﬁes $(\alpha,\rho)$-RDP if
\begin{align*}
    \rho \leq D_\alpha\Big(\gN(0,\sigma^2) \, \big\Vert \, (1-q)\gN(0,\sigma^2)+q\gN(1,\sigma^2)\Big) \\
    \text{and} \quad \rho \leq D_\alpha\Big((1-q)\gN(0,\sigma^2)+q\gN(1,\sigma^2) \, \big\Vert \, \gN(0,\sigma^2) \Big)
\end{align*}
\end{theorem}

Theorem~\ref{theorem:one-dim} reduce the problem of proving the RDP bound for $\mathrm{SG}_{q,\sigma}$ to a simple special case of a mixture of one-dimensional Gaussians.

\begin{theorem}\cite{mironov2019r}.
\label{th:mix_simple} Let $\mu_0$ denote the $\mathrm{pdf}$ of $\mathcal{N}(0,\sigma^2)$, 
 $\mu_1$ denote the $\mathrm{pdf}$ of $\mathcal{N}(1,\sigma^2)$, and let $\mu$ be the mixture of two Gaussians $\mu= (1-q)\mu_0 + q\mu_1$.
    Let $\mathrm{SG}_{q,\sigma}$ be the Sampled Gaussian mechanism for some function $f$ and under the assumption $\Delta_2 f \leq 1$ for any adjacent $E, E' \in \mathcal{E}$. Then $\mathrm{SG}_{q,\sigma}$ satisfies $(\alpha,\rho)$-RDP if
    \begin{equation}
    \label{eq:logmax}
    \rho \leq \frac{1}{\alpha-1} \log{(\max \{A_{\alpha},B_{\alpha}\})}
\end{equation}
where $A_{\alpha} \overset{\Delta}{=} \mathbb{E}_{z\sim \mu_0} [\left( \mu(z)/\mu_0(z)\right)^\alpha] $
  and $ B_{\alpha} \overset{\Delta}{=} \mathbb{E}_{z\sim \mu} [\left( \mu_0(z)/\mu(z)\right)^\alpha]$ 

\end{theorem}

Theorem~\ref{th:mix_simple} states that applying $\mathrm{SGM}$ to a function of sensitivity (\equationautorefname~\ref{def:sensitivity}) at most 1 (which also holds for larger values without loss of generality) satisfies $(\alpha,\rho)$-RDP if  $\rho \leq \frac{1}{\alpha-1} \log(\max \{A_{\alpha},B_{\alpha}\} )$. Thus, analyzing RDP properties of SGM is equivalent to upper bounding $A_{\alpha}$ and $B_{\alpha}$.

From Corollary 7 in~\citet{mironov2019r}, $A_{\alpha}\geq B_{\alpha}$ for any $\alpha \geq 1$. Therefore, we can reformulate \ref{eq:logmax} as
\begin{equation}
    \label{eq:xi}
    \rho \leq \xi_{\mathcal{N}}(\alpha|q) \coloneqq \frac{1}{\alpha-1} \log A_{\alpha}
\end{equation}

To compute $A_{\alpha}$, we use the numerically stable computation approach proposed in Section 3.3 of \citet{mironov2019r}. The specific approach used depends on whether $\alpha$ is expressed as an integer or a real value.

\begin{theorem}[Composability~\cite{mironov2017renyi}]
\label{th:comp}
Suppose that a mechanism $\mathcal{M}$ consists of a sequence of adaptive mechanisms $\mathcal{M}_1, \dots, \mathcal{M}_k$ where $\mathcal{M}_i: \prod_{j=1}^{i-1} \mathcal{R}_j \times \mathcal{E} \rightarrow \mathcal{R}_i$. If all mechanisms in the sequence are $(\alpha,\rho)$-RDP, then the composition of the sequence is $(\alpha,k\rho)$-RDP.
\end{theorem}
In particular, Theorem~\ref{th:comp} holds 
 when the mechanisms themselves are chosen based on the (public) output of the
previous mechanisms. By Theorem~\ref{th:comp}, it suffices to compute $\xi_{\mathcal{N}}(\alpha|q)$ at each step and sum them up to bound the overall RDP privacy budget of an iterative mechanism composed of single DP mechanisms at each step.

\begin{theorem}[Conversion from RDP to DP~\cite{balle2020hypothesis}]
\label{th:conv}
If a mechanism $\mathcal{M}$ is $(\alpha,\rho)$-RDP then it is 
$((\rho + \log((\alpha-1)/\alpha) - (\log \delta + \log \alpha)/(\alpha -1),\delta)$-DP for any $0<\delta<1$.
\end{theorem}

\begin{theorem}[Privacy of FedLAP-DP]
For any $0<\delta<1$ and $\alpha \geq 1$, FedLAP-DP is $(\varepsilon,\delta)$-DP, with 
\begin{equation}
\begin{split}
    \varepsilon = \min_{\alpha}& \Big(M \cdot \xi_\gN(\alpha|q_1) + M(T -1) \cdot \xi_\gN(\alpha|q_2) \\ 
    &+ \log((\alpha-1)/\alpha) - (\log \delta + \log \alpha)/(\alpha -1)\Big)    
\end{split}
\end{equation}

Here, $\xi_{\mathcal{N}}(\alpha|q)$ is defined in Eq.~\ref{eq:xi}, $q_1=\frac{C\cdot \mathbb{B}}{\min_k |\gD_k|}$, 
 $q_2=\frac{\mathbb{B}}{\min_k |\gD_k|}$, $M$ is the number of federated rounds, $T$ is the total number of local updates (i.e., total accesses to local private data) per federated round, $C$ is the probability of selecting any client per federated round, $\mathbb{B}$ is the local batch size, and $|\gD_k|$ denotes the local dataset size.
\end{theorem}

The proof follows from Theorems~\ref{th:mix_simple}, \ref{th:comp},\ref{th:conv} and the fact that a record is sampled in the very first SGD iteration of every round if two conditions are met.  First, the corresponding client must be selected, which occurs with a probability of $C$. Second, the locally sampled batch at that client must contain the record, which has a probability of at most $\frac{\mathbb{B}}{\min_k |\mathcal{D}_k|}$. However, the adaptive composition of consecutive SGD iterations are considered where the output of a single iteration depends on the output of the previous iterations. Therefore, the sampling probability for the first batch is $q_1=\frac{C\cdot \mathbb{B}}{\min_k |\gD_k|}$, while the sampling probability for every subsequent SGD iteration within the same round is at most $q_2=\frac{\mathbb{B}}{\min_k |\gD_k|}$ \textit{conditioned} on the result of the first iteration~\cite{kerkouche2021privacy}.

%% file: articles/experiments.tex
\begin{table*}[tb]
\small
\aboverulesep=0.1ex
\belowrulesep=0.1ex
\begin{tabularx}{\linewidth}{@{}X|Y|YYY|YYY@{}}
\toprule
 & PSG$^\dag$~\cite{chenprivate} & DP-FedAvg & DP-FedProx & Ours & DP-FedAvg & DP-FedProx & Ours \\
 \midrule
  \multicolumn{1}{c|}{$\varepsilon$} & 32\ & \multicolumn{3}{c|}{2.79}& \multicolumn{3}{c}{10.18} \\ \midrule
MNIST &  88.34±0.8 & 45.25±6.9 & 54.58±4.9 & \textbf{60.72±1.3} & 86.99±0.5 & \textbf{88.75±0.5} & 87.77±0.8 \\
FMNIST & 67.91±0.3 & 50.11±4.2 & 54.57±2.9 & \textbf{59.85±1.5} & 72.78±1.3 & 71.67±2.2 & \textbf{73.00±0.7} \\
CIFAR-10 & 34.58±0.4  & 17.11±0.7 & 19.40±0.7 & \textbf{21.42±1.4} & 31.15±0.4 & 35.04±1.1 & \textbf{36.09±0.5} \\ \bottomrule
\end{tabularx}
\caption{
Utility and Privacy budgets at varying privacy regimes. The high privacy regime with $\varepsilon=2.79$ corresponds to the first communication round, while a privacy level of $\varepsilon=10.18$ represents the commonly considered point ($\varepsilon$=10) in private learning literature. PSG$^\dag$ corresponds to a one-shot centralized setting and is reproduced from the official code with the default configuration that yields $\varepsilon=32$ in federated settings.
} \vspace{-0.5cm}
\end{table*}

\begin{figure*}[tb]
    \centering
    \includegraphics[width=\linewidth]{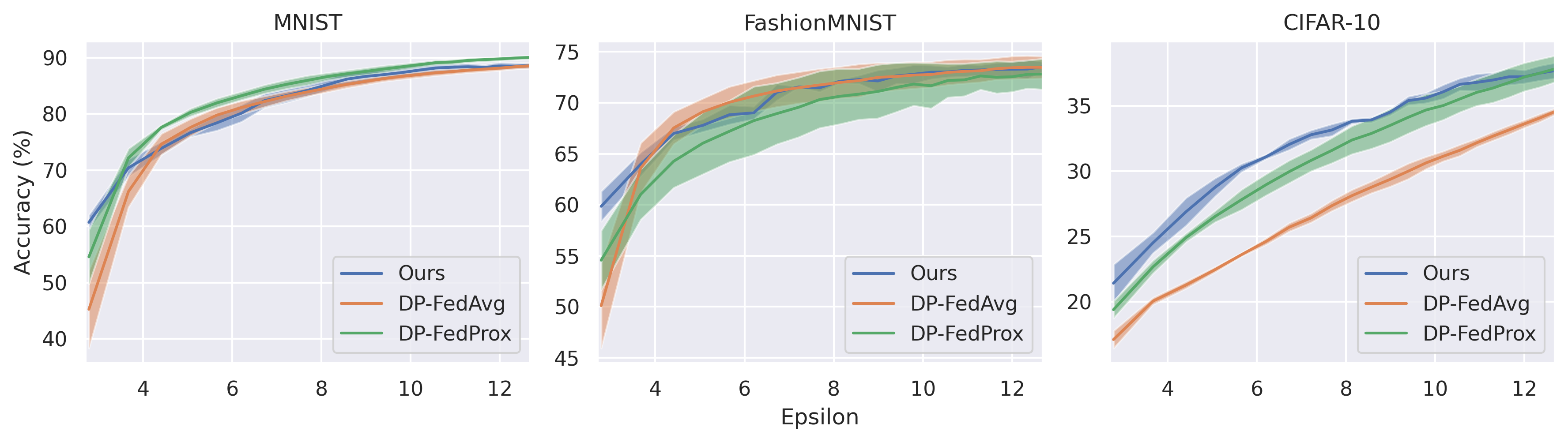}
    \caption{Privacy-utility trade-off with $\delta=10^{-5}$. A \textit{smaller} value of $\varepsilon$ (x-axis) indicates a \textit{stronger} privacy guarantee. Evaluation is conducted at each communication round.}
    \label{fig:dp}
\end{figure*}

\section{Experiments}
\label{sec:exp}

\subsection{Setup}
\label{ssec:exp-setup}

\myparagraph{Overview} We consider a standard classification task by training federated ConvNets~\citep{lecun2010convolutional} on three benchmark datasets: MNIST~\citep{lecun1998gradient}, FashionMNIST~\citep{xiao2017fashion}, and CIFAR-10~\citep{krizhevsky2009learning}. Our study focuses on a non-IID setting where five clients possess disjoint class sets, meaning each client holds two unique classes. This scenario is typically considered challenging~\citep{hsu2019measuring} and mirrors the cross-silo setting~\citep{kairouz2021advances} where all clients participate in every training round while maintaining a relatively large amount of data, yet exhibiting statistical divergence (e.g., envision the practical scenario for collaborations among hospitals). 
Our method employs a learning rate of $100$ for updating synthetic images and $0.1$ with cosine decay for model updates. We set by default $(R_i, R_l, R_b, r) = (4, 2, 10, 1.5)$ and $(1, 0, 5, 10)$ for DP and non-DP training, respectively. Additionally, we include a weight of 0.1 for Mean Squared Error (MSE) regularization in our method (\equationautorefname~\ref{eq:local_approx}). To prevent infinite loops caused by the neighborhood search, we upper bound the while loops in Algorithm~\ref{alg:nondp-client} by $5$ iterations. We follow FL benchmarks~\citep{mcmahan2017communication,reddi2021adaptive} and the official codes for training the baselines. %
All experiments are repeated over three random seeds. In this work, we only consider a balanced setting where every client owns the same amount of training samples. We acknowledge that minor sample imbalances can be managed by existing federated algorithms via modifying the aggregation weights, for example, $N_k/N$ as indicated in \equationautorefname~\ref{eq:recover_global}. However, severe imbalances present a significant challenge that might affect the efficacy of both our method and gradient-sharing techniques, and addressing this issue falls beyond the purview of the current study.

\vspace{5pt}\myparagraph{Architecture} We provide the details of the federated ConvNet used in our paper. The network consists of three convolutional layers, followed by two fully-connected layers. ReLU activation functions are applied between each layer. Each convolution layer, except for the input layer, is composed of $128$ (input channels) and $128$ (output channels) with $3\times3$ filters. Following prior work~\cite{wang2022progfed, reddi2021adaptive, zhao2023dataset, cazenavette2022dataset}, we attach Group Normalization~\cite{wu2018group} before the activation functions to stabilize federated training. For classification, the network utilizes a global average pooling layer to extract features, which are then fed into the final classification layer for prediction. The entire network contains a total of 317,706 floating-point parameters. The model details are listed in the appendix.

\subsection{Data Heterogeneity}
\label{ssec:exp-non-iid}

We first demonstrate the effectiveness of FedLAP over various baselines on benchmark datasets in a non-IID setting. Our method assigns 50 images to each class, resulting in comparable communication costs to the baselines.
The baselines include: \textbf{DSC}~\citep{zhao2021dataset}, the dataset distillation method considering centralized one-shot distillation; \textbf{FedSGD}~\citep{mcmahan2017communication}, that transmits every single batch gradient to prevent potential model drifting; \textbf{FedAvg}~\citep{mcmahan2017communication}, the most representative FL method; \textbf{FedProx}~\citep{li2020federated}, \textbf{SCAFFOLD}~\citep{karimireddy2020scaffold}, state-of-the-art federated optimization for non-IID distributions, and \textbf{FedDM}~\citep{xiong2023feddm}, a concurrent work that shares a similar idea but without considering approximation quality. Note that DSC operates in a (one-shot) centralized setting, SCAFFOLD incurs double the communication costs compared to the others by design, and FedDM requires class-wise optimization.
As depicted in \tableautorefname~\ref{tab:perf_comparison}, our method surpasses DSC and FedSGD, highlighting the benefits of multi-round training on the server and client sides, respectively. Moreover, our method presents superior performance over state-of-the-art optimization methods, validating the strength of optimizing from a global view. We also plot model utility over training rounds in \figureautorefname~\ref{fig:non-iid-results}, where our method consistently exhibits the fastest convergence across three datasets. In other words, our methods consume fewer costs to achieve the same or better performance level and more communication efficiency.

\subsection{Privacy Protection}
\label{ssec:exp_dp}

We evaluate the trade-off between utility and privacy costs $\varepsilon$ on benchmark datasets against two state-of-the-art methods, DP-FedAvg (the local DP version in~\citet{truex2019hybrid}) and DP-FedProx. Note that FedDM~\citep{xiong2023feddm} is incomparable since it considers class-wise optimization, introducing additional privacy risks and a distinct privacy notion.
Our method assigns 10 images per class and is evaluated under the worst-case scenario, i.e., we assume the maximum of 5 while loops is always reached (\sectionautorefname~\ref{ssec:exp-setup}) for the $\varepsilon$ computation, despite the potential early termination (and thus smaller $\varepsilon$) caused by the radius $r$ (\equationautorefname~\ref{eq:local_approx_cst}). 
To ensure a fair and transparent comparison, we require our method to access the same amount of private data as the baselines in every communication round and consider a noise scale $\sigma=1$ for all approaches. %
\figureautorefname~\ref{fig:dp} demonstrates that our framework generally exhibits superior performance, notably with smaller $\varepsilon$ and more complex dataset such as CIFAR-10. This superiority is further quantified in \tableautorefname~\ref{tab:perf_comparison} under two typical privacy budgets of $2.79$ and $10.18$. Moreover, when compared to the private one-shot dataset condensation method (\textbf{PSG}~\citep{chenprivate}), our approach presents a better privacy-utility trade-off, effectively leveraging the benefits of multi-round training in the challenging federated setting.

\subsection{Ablation Study}
\label{ssec:ablation}

\begin{table}[t]
    \begin{tabularx}{\linewidth}{@{}YYYY@{}}
        \toprule
        Min & Max & Median & Fixed \\ \midrule
        71.90 & 72.39 & 72.33 & 71.26 \\ \bottomrule
    \end{tabularx}
    \caption{Performance comparison between radius selection strategies. } \vspace{-0.5cm}
    \label{table:strategy}
\end{table}

\myparagraph{Radius Selection} As in \sectionautorefname~\ref{sec:method}, we assess various server optimization radius selection strategies: \emph{Fixed}, \emph{Max}, \emph{Median}, and \emph{Min}. The \emph{Fixed} strategy employs a static length of 100 iterations regardless of quality, \emph{Max} pursues swift optimization with the largest radius, \emph{Median} moderates by adhering to the majority, and \emph{Min}—used in all experiments—targets the safest region agreed by all synthetic image sets. \tableautorefname~\ref{table:strategy} shows that strategies mindful of approximation quality surpass the fixed approach. Detailed analysis in \sectionautorefname~\ref{ssec:appendix-radius-selection} reveals that aggressive strategies yield inferior intermediate performance, unsuitable for federated applications needing satisfactory intermediate results. Among the strategies, \emph{Min} proves optimal.

\begin{figure*}[t]
    \centering
    \includegraphics[width=\linewidth]{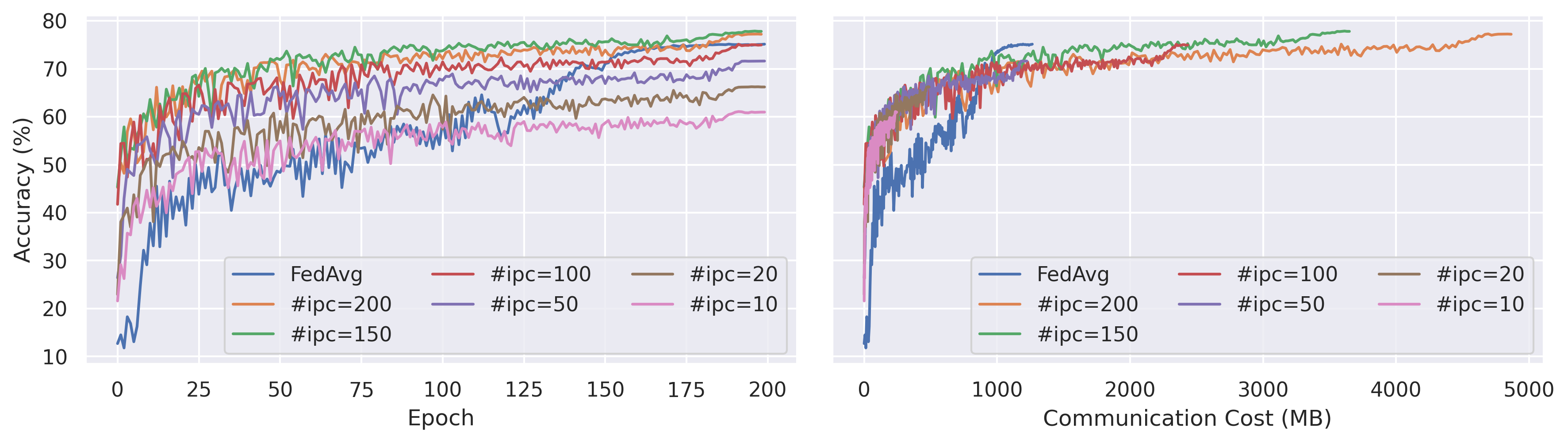}
    \caption{Ablation study on the number of images per class (\#ipc).}
    \label{fig:ablation_ipc}
\end{figure*}

\myparagraph{Size of Synthetic Datasets} We investigate the impact of synthetic dataset size on approximation. In general, higher numbers of synthetic samples submitted by clients lead to greater information communication. To further explore this concept, we conducted experiments on CIFAR-10, building upon the previous experiment (shown in \figureautorefname~\ref{fig:non-iid-results}) by adding five additional settings in which we assigned 10, 20, 100, 150, and 200 images to each class (referred to as "image per class" or \#ipc). Our results, presented in \figureautorefname~\ref{fig:ablation_ipc}, demonstrate that our method performs best when assigning 200 images, supporting the hypothesis that more synthetic samples convey more information. Additionally, our method produces superior outcomes regardless of \#ipc when communication costs are restricted, making it advantageous for resource-constrained devices.

\begin{table}[tb]
\begin{tabular}{@{}lcccc@{}}
\toprule
 & 5\% & 10\% & 50\% & 100\% \\ \midrule
FedAvg & 55.16 (-20.04) & 61.43 (-13.77) & 71.34 (-3.86) & 75.20 \\
Ours & 56.34 (-15.57) & 62.76 (-9.15) & 68.97 (-2.94) & 71.91 \\ \bottomrule
\end{tabular}
\caption{Impact of different training set sizes (5\%, 10\%, 50\%, and 100\% of the original dataset): Parenthesized values indicate the reduction in performance when compared to utilizing the full training set (100\%)} \vspace{-0.5cm}
\label{table:size}
\end{table}

\myparagraph{Size of Training Sets} 
In this experiment, we show that given the same amount of optimization steps, our method can approximate the information more faithfully as the number of training examples that each client can access decreases.
\tableautorefname~\ref{table:size} presents the impact of varying training set sizes. We sub-sample 5\%, 10\%, 50\%, and 100\% of samples from the original CIFAR-10 dataset and repeat the same experiments as in \tableautorefname~\ref{tab:perf_comparison}. In other words, each client has 500, 1000, and 5000 images, respectively. It is observed that the performance of both methods degrades as the number of training samples decreases. Our method demonstrates greater resilience to reductions in training sample size than FedAvg by exhibiting less performance degradation when compared to using the full training set. A potential explanation is that a smaller sample size may be simpler to approximate given the same number of optimization steps.

\begin{figure*}[t]
    \centering
    \includegraphics[trim=0cm 2.2cm 0cm 0cm,clip,width=\linewidth]{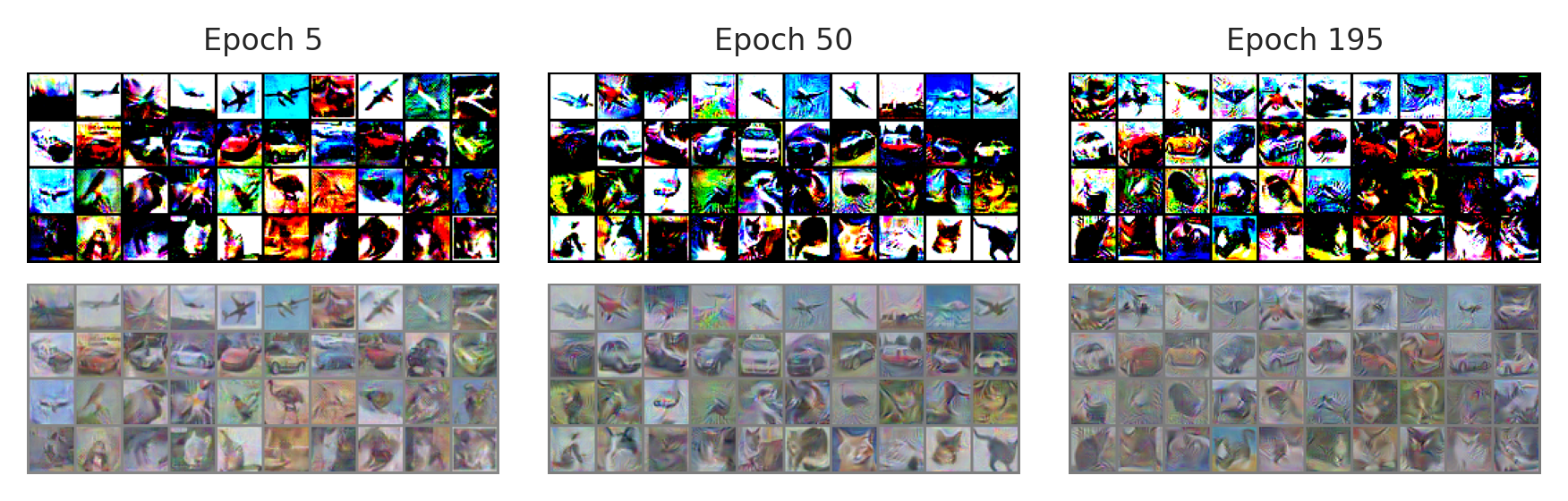}
    \vskip -0.1in
    \caption{Visualization of synthetic images at epochs 5, 50, and 196 in the non-private CIFAR-10 experiment. The pixel values are clipped to the range $[0, 1]$. Each row corresponds to airplane, automobile, bird, and cat, respectively.}
    \label{fig:vis}
\end{figure*}

\begin{figure*}[t]
    \centering
    \includegraphics[trim=0cm 2.2cm 0cm 0cm,clip,width=\linewidth]{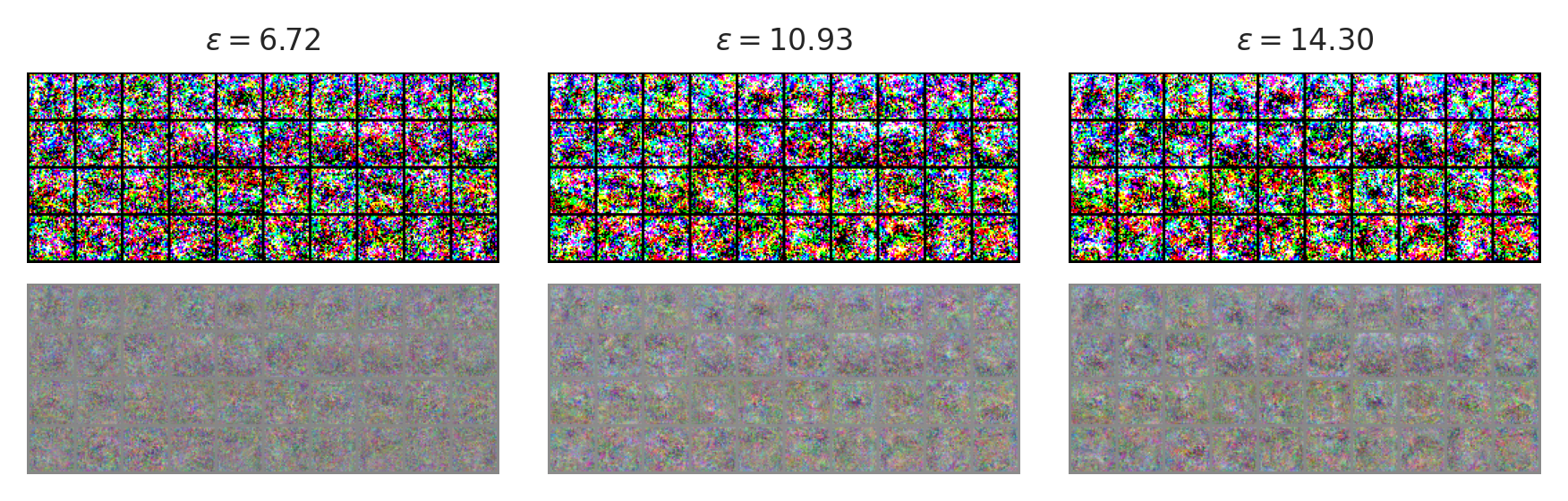}
    \vskip -0.1in
    \caption{Visualization of synthetic images for $\varepsilon=6.72\text{, }10.93\text{, }\text{and }14.30$ in the privacy-preserving CIFAR-10 experiment. The pixel values are clipped to the range $[0, 1]$. Each row corresponds to airplane, automobile, bird, and cat, respectively.}
    \label{fig:vis-dp}
\end{figure*}

\subsection{Qualitative Results}
\label{ssec:qualitative}
We visualize the synthetic images generated at different training stages in both DP and non-DP settings. Specifically, we consider the same setting as in \sectionautorefname~\ref{ssec:exp-non-iid} and \sectionautorefname~\ref{ssec:exp_dp} on CIFAR-10. We randomly sample synthetic images associated with classes \emph{airplane}, \emph{automobile}, \emph{bird}, and \emph{cat} at epochs 5, 50, and 196. \figureautorefname~\ref{fig:vis} shows that the images synthesized by our approach look drastically different from real images even in non-DP settings. Most of the details have been obfuscated to carry essential gradient information. Notably, a recent work~\citep{dong2022privacy} arguably suggests that dataset distillation may naturally introduce privacy protection, making models more robust than the ones trained by plain gradients. 

On the other hand, \figureautorefname~\ref{fig:vis-dp} visualizes the synthetic images generated under the DP protection of $\varepsilon=\{6.72, 10.93, 14.30\}$. It is observed that even with the loosest privacy budgets ($\varepsilon=14.30$), the images tend to look like random noise and significantly obfuscate most of the visual contents compared to \figureautorefname~\ref{fig:vis}. The protection gets further stronger as the privacy budget decreases. These visualizations verify the effectiveness of our approach in introducing record-level DP and might offer a new tool to visually examine the privacy protection of the communicated information.

\subsection{Privacy Auditing}
\label{ssec:audit}

Despite the theoretical privacy protection introduced by DP, we examine the empirical privacy protection by conducting membership inference attacks~\citep{kumar2020mlprivacy} on both our method and FedAvg in DP and non-DP scenarios, respectively. In particular, we consider two attack settings. We first divide the testing set (i.e., non-members) into two disjoint partitions and use one of them for training attack models while using the other as the audit dataset. In the first setting, we initially collect the loss values of the examples in a black-box manner, which are later used to train a neural network to distinguish whether a given example belongs to the training set. On the other hand, the second setting adopts the gradient norms of the examples, providing more detailed information for membership inference attacks. The access of gradient norms demands a white-box setting, i.e., assuming the model parameters are accessible.

We present the Receiver Operating Characteristic (ROC) curves and the Area Under the Curve (AUC) scores for both scenarios in Figures \ref{fig:mia-black-box} and \ref{fig:mia-white-box}. The diagonal lines present the baseline performance of random guesses, which are characterized by AUC scores of 0.5. Methods that perform closer to random guesses provide better privacy protection since the adversaries hardly infer information from the victim models. It is observed that methods without DP protection are vulnerable to membership inference attacks. Notably, our methods are more robust than gradient-sharing schemes, even without DP protection. The resilience may stem from the approximation process that further limits information leakage. In contrast, the methods with DP protection consistently produce AUC scores around 0.5 regardless of attack settings, verifying the effectiveness of DP protection.

%% file: articles/discussion.tex
\section{Discussion}
\label{sec:discussion}

\myparagraph{Future Directions} While the primary contribution of FedLAP-DP lies in utilizing local approximation for global optimization, we demonstrate in the appendix that its performance can be further enhanced by improving the quality of the approximation. Moreover, ongoing research in synthetic data generation~\citep{zhao2021dataset, zhao2023dataset, cazenavette2022dataset} represents a potential avenue for future work, which could potentially benefit our formulation. The potential directions include explicitly matching training trajectories~\citep {cazenavette2022dataset} or leveraging off-shelf foundation models~\citep{cazenavette2023generalizing}. Moreover, we also observe that given the same amount of optimization steps, the performance of our method may decrease when the number of classes increases on clients. For instance, the performance on CIFAR-10 in an IID scenario drops from 71.91 to 62.52 while FedAvg remains at a similar level of 73.5. This suggests that if there is no performance degradation caused by data heterogeneity, our method loses information during image synthesis. Future works that improve the efficiency of approximation could further bridge such gap and enable more efficient federated learning under extremely non-IID scenarios.

\myparagraph{Computation Overhead} Our method suggests an alternative to current research, trading computation for improved performance and communication costs incurred by slow convergence and biased optimization. We empirically measured the computation time needed for a communication round by a client. We observed an increase from 0.5 minutes (FedAvg) to 2.5 minutes (FedLAP) using one NVIDIA Titan X. Despite the increase, the computation time is still manageable in cross-silo environments, where participants are deemed to have more computation power. A thorough analysis can be found in \sectionautorefname~\ref{sec:appendix-computation}. We anticipate this work will motivate the community to further explore the trade-off between computation and communication resource demands beyond local epochs, as we have demonstrated in \figureautorefname~\ref{fig:ablation_ipc}.

\begin{figure*}[t]
\begin{tabular}{cc}
 \includegraphics[width=0.49\linewidth]{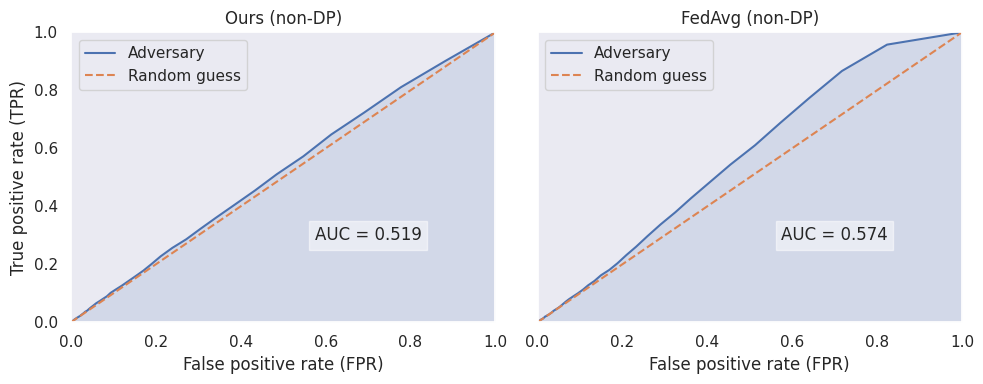} & \includegraphics[width=0.49\linewidth]{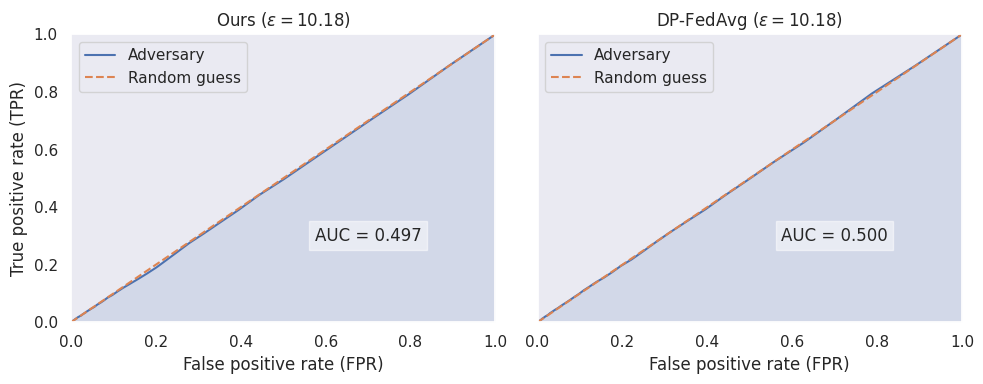} \\
 (a) non-DP & (b) DP ($\varepsilon=10.18$)
\end{tabular}
\caption{Black-box membership inference attacks based on loss values targeting FedLAP (ours) and FedAvg under non-DP and DP ($\varepsilon=10.18$) settings.}
\label{fig:mia-black-box}
\end{figure*}

\begin{figure*}[h]
\begin{tabular}{cc}
 \includegraphics[width=0.49\linewidth]{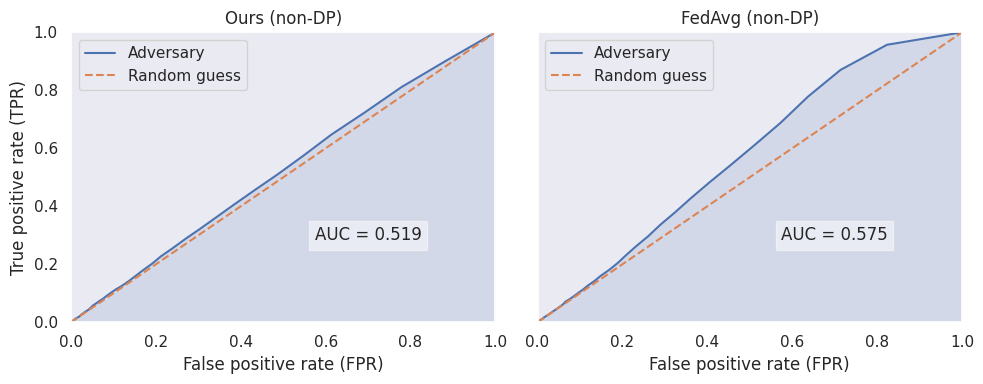} & \includegraphics[width=0.49\linewidth]{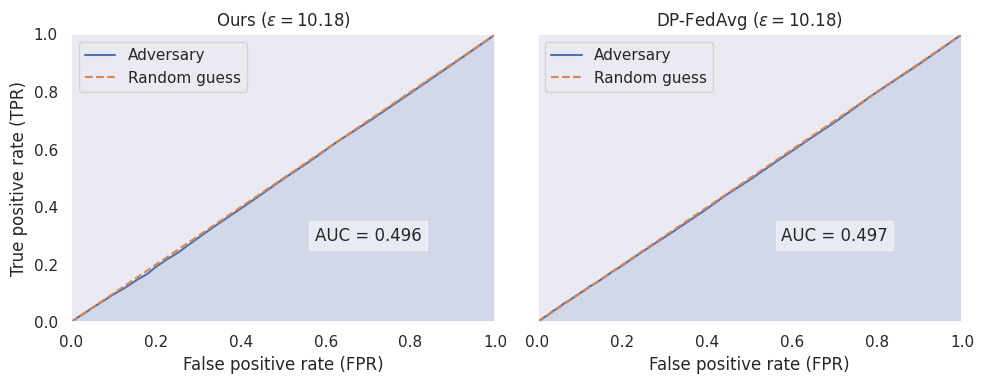} \\
 (a) non-DP & (b) DP ($\varepsilon=10.18$)
\end{tabular}
\caption{White-box membership inference attacks based on gradient norms targeting FedLAP (ours) and FedAvg under non-DP and DP ($\varepsilon=10.18$) settings.}
\label{fig:mia-white-box}
\end{figure*}

\myparagraph{(Visual) Privacy} \citet{dong2022privacy} were among the first to show that data distillation may naturally offer superior privacy protections compared to traditional gradient information in centralized settings. In \sectionautorefname~\ref{ssec:audit} of our work, we extend this concept into federated environments, although our synthetic images are not intended to resemble realistic or class-specific content. Despite these advances, a comprehensive analysis of existing dataset distillation approaches in terms of privacy remains pending. This necessitates further exploration in diverse settings, including but not limited to, defense against reconstruction attacks and membership inference, as well as an in-depth privacy comparison with non-DP gradient-sharing techniques. Beyond the realm of membership privacy, safeguarding visual privacy—the discernible content within images—remains a complex issue~\citep{padilla2015visual}. While several strategies like adversarial entropy maximization~\citep{wang2021infoscrub} and image masking~\citep{paruchuri2009video} have been explored, finding the right balance between utility and privacy varies with the use case and presents opportunities for enhancement.

%% file: articles/conclusion.tex
\section{Conclusion}
\label{sec:conclusion}

In conclusion, this work introduces FedLAP-DP, a novel approach for privacy-preserving federated learning. FedLAP-DP utilizes synthetic data to approximate local loss landscapes within calibrated trust regions, %
effectively debiasing the optimization on the server. Moreover, our method seamlessly integrates record-level differential privacy, ensuring strict privacy protection for individual data records. Extensive experimental results demonstrate that FedLAP-DP outperforms gradient-sharing approaches in terms of faster convergence on highly-skewed data splits and reliable utility under differential privacy settings. We further explore the critical role of radius selection, the influence of synthetic dataset size, open directions, and potential enhancements to our work. 
Overall, FedLAP-DP presents a promising approach for privacy-preserving federated learning, addressing the challenges of convergence stability and privacy protection in non-IID scenarios.

%% file: articles/appendix.tex
\begin{figure*}[tbp]
    \centering
    \begin{lstlisting}[basicstyle=\fontsize{8}{9}\selectfont]
    ConvNet(
      (features): Sequential(
        (0): Conv2d(3, 128, kernel_size=(3, 3), stride=(1, 1), padding=(1, 1))
        (1): GroupNorm(128, 128, eps=1e-05, affine=True)
        (2): ReLU(inplace=True)
        (3): AvgPool2d(kernel_size=2, stride=2, padding=0)
        (4): Conv2d(128, 128, kernel_size=(3, 3), stride=(1, 1), padding=(1, 1))
        (5): GroupNorm(128, 128, eps=1e-05, affine=True)
        (6): ReLU(inplace=True)
        (7): AvgPool2d(kernel_size=2, stride=2, padding=0)
        (8): Conv2d(128, 128, kernel_size=(3, 3), stride=(1, 1), padding=(1, 1))
        (9): GroupNorm(128, 128, eps=1e-05, affine=True)
        (10): ReLU(inplace=True)
        (11): AvgPool2d(kernel_size=2, stride=2, padding=0)
      )
      (classifier): Linear(in_features=2048, out_features=10, bias=True)
    )
    \end{lstlisting} 
    \caption{Architecture of ConvNets used in the federated experiments.}
\end{figure*}

\section{Additional Analysis}
\begin{figure*}[tbp]
    \centering
    \includegraphics[width=\linewidth]{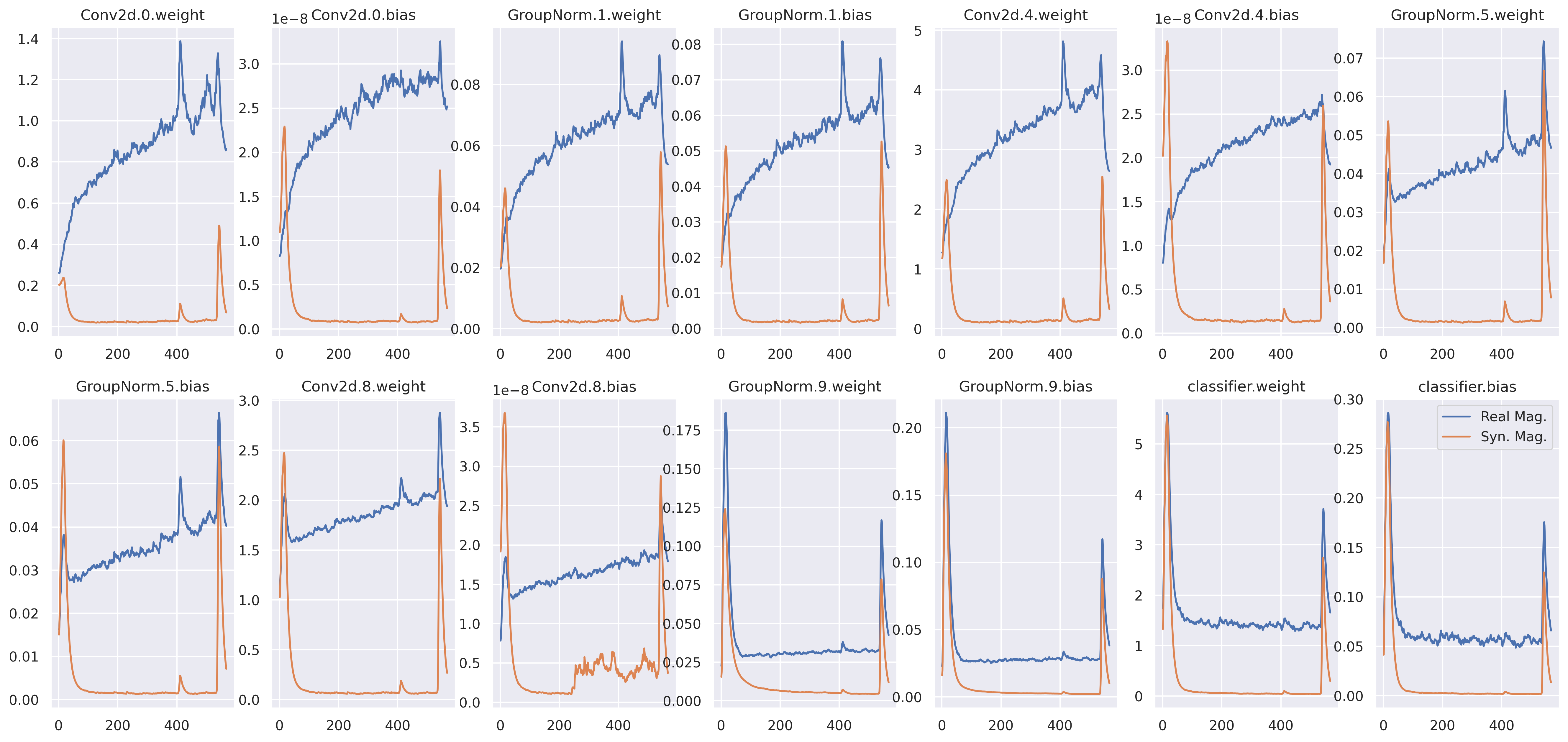}
    \caption{Discrepancy in gradient magnitudes between real and synthetic data. The noticeable difference in magnitudes during training highlights the limitation of solely regulating gradient directions, as optimizing without magnitude information introduces training instability.}
    \label{fig:appendix-mag-mismatch}
\end{figure*}

\begin{figure}[tb]
    \centering
    \includegraphics[width=\linewidth]{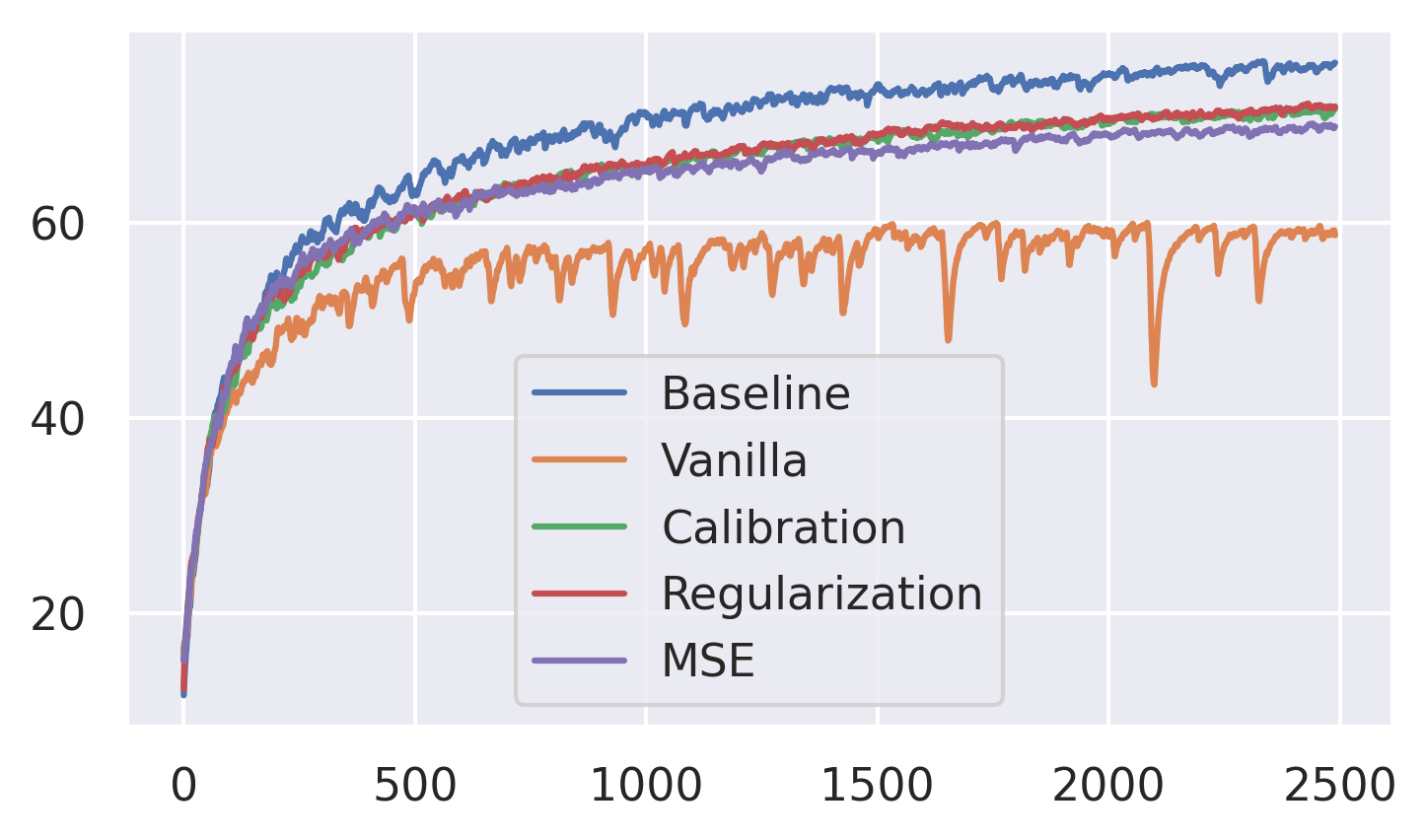}
    \caption{Performance comparison of fitting one batch. \textbf{Baseline}: FedSGD with one client. \textbf{Vanilla}: our method with cosine similarity. \textbf{Calibration}: our method with cosine similarity and magnitude calibration. \textbf{Regularization}: our method with cosine similarity and MSE regularization (i.e., \equationautorefname~\ref{eq:local-loss}). \textbf{MSE}: gradient matching by solely measuring mean square errors (\equationautorefname~\ref{eq:appendix-matching-mse}).}
    \label{fig:appendix-fit-batch-reg}
\end{figure}

\subsection{Matching Criteria}
\label{ssec:appendix-matching-criteria}
We analyze our design choices by using our method to fit gradients computed on a single batch. We simplify the learning task by employing only one client that contains all training data, resembling centralized training or FedSGD with a single client. In this setup, the client immediately communicates the gradients to the central server after computing them on a single batch. This scenario can be seen as a trivial federated learning task, as it does not involve any model drifting (non-IID) or communication budget constraints. It is worth noting that the baseline performance in this setting is an \emph{ideal} case that does not apply in any practical FL use cases (or out of scope of federated learning).

\paragraph{Gradient Magnitudes} While previous research~\cite{zhao2021dataset, zhao2021dataset2, he2021cossgd} suggests that gradient directions are more crucial than magnitudes (\equationautorefname~\ref{eq:local-loss}), our study demonstrates that as training progresses, the magnitudes of synthetic gradients (i.e., gradients obtained from synthetic images) can differ significantly from real gradients. In \figureautorefname~\ref{fig:appendix-mag-mismatch}, we display the gradient magnitudes of each layer in a ConvNet. Our findings indicate that even with only 500 iterations, the magnitudes of synthetic gradients (orange) noticeably deviate from the real ones (blue), causing unnecessary instability during training.

\myparagraph{Post-hoc Magnitude Calibration} To further validate the issue, we implement a post-hoc magnitude calibration, called \emph{Calibration} in \figureautorefname~\ref{fig:appendix-fit-batch-reg}. It calibrates the gradients obtained from the synthetic images on the server. Specifically, the clients send the layer-wise magnitudes of real gradients $\Vert \nabla_\mathbf{w} \Ls(\mathbf{w},\gD_k)\Vert$ to the server, followed by a transformation on the server:
\begin{equation}
\label{eq:appendix_calibration}
    \frac{\nabla_\mathbf{w} \Ls(\mathbf{w},\gS_k)}{\Vert\nabla_\mathbf{w} \Ls(\mathbf{w},\gS_k)\Vert} \Vert \nabla_\mathbf{w} \Ls(\mathbf{w},\gD_k)\Vert.
\end{equation}
In \figureautorefname~\ref{fig:appendix-fit-batch-reg}, We observe that the synthetic images with the magnitude calibration successfully and continuously improve over the one without the calibration (Vanilla). It implies that even with the same cosine similarity, inaccurate gradient magnitudes could dramatically fail the training. It is worth noting that the performance gap between the baseline and our method in this experiment does not apply to federated learning since FL models suffer from non-IID problems induced by multiple steps and clients, the gradient-sharing schemes, such as FedAvg, overly approximate the update signal, further enhancing the problem. Meanwhile, it also suggests an opportunity to improve our method if future work can further bridge the gap.

\myparagraph{Mean Square Error (MSE) Regularization} Although calibration can improve performance, it is not suitable for federated learning due to two reasons. Firstly, when the synthetic images $\gS_k$ from different clients are merged into a dataset for server optimization, it remains unclear how to apply \equationautorefname~\ref{eq:local-loss} to the averaged gradients of the merged dataset $\gS$, especially when multiple-step optimization is involved. Secondly, transmitting magnitudes can pose privacy risks. Instead of explicit calibration, we propose using MSE regularization (\equationautorefname~\ref{eq:local-loss}, termed \emph{Regularization} in \figureautorefname~\ref{fig:appendix-fit-batch-reg}) to limit potential magnitude deviations while focusing on the directions. As depicted in \figureautorefname~\ref{fig:appendix-fit-batch-reg}, our proposed method remains close to the calibration method, suggesting that regularization prevents mismatch. 

Moreover, we present an additional implementation with an MSE matching criterion (termed \emph{MSE} in \figureautorefname~\ref{fig:appendix-fit-batch-reg})). It solely matches the mean square error distance between $\Ls(\mathbf{w},\gD_k)$ and $\Ls(\mathbf{w},\gS_k)$ regardless of gradient directional information. That is,

\begin{equation}
    \label{eq:appendix-matching-mse}
    \Vert \nabla_{\mathbf{w}^{(l)}} \Ls(\mathbf{w}^{(l)},\gD_k
        ) - \nabla_{\mathbf{w}^{(l)}} \Ls (\mathbf{w}^{(l)},\gS_k) \Vert_2^2 
\end{equation}

Despite the improvement over the vanilla method, it still falls behind \emph{Calibration} and \emph{Regularization}, highlighting the importance of directional information and justifying our design choice.

\begin{figure}[tb]
    \centering
    \includegraphics[width=\linewidth]{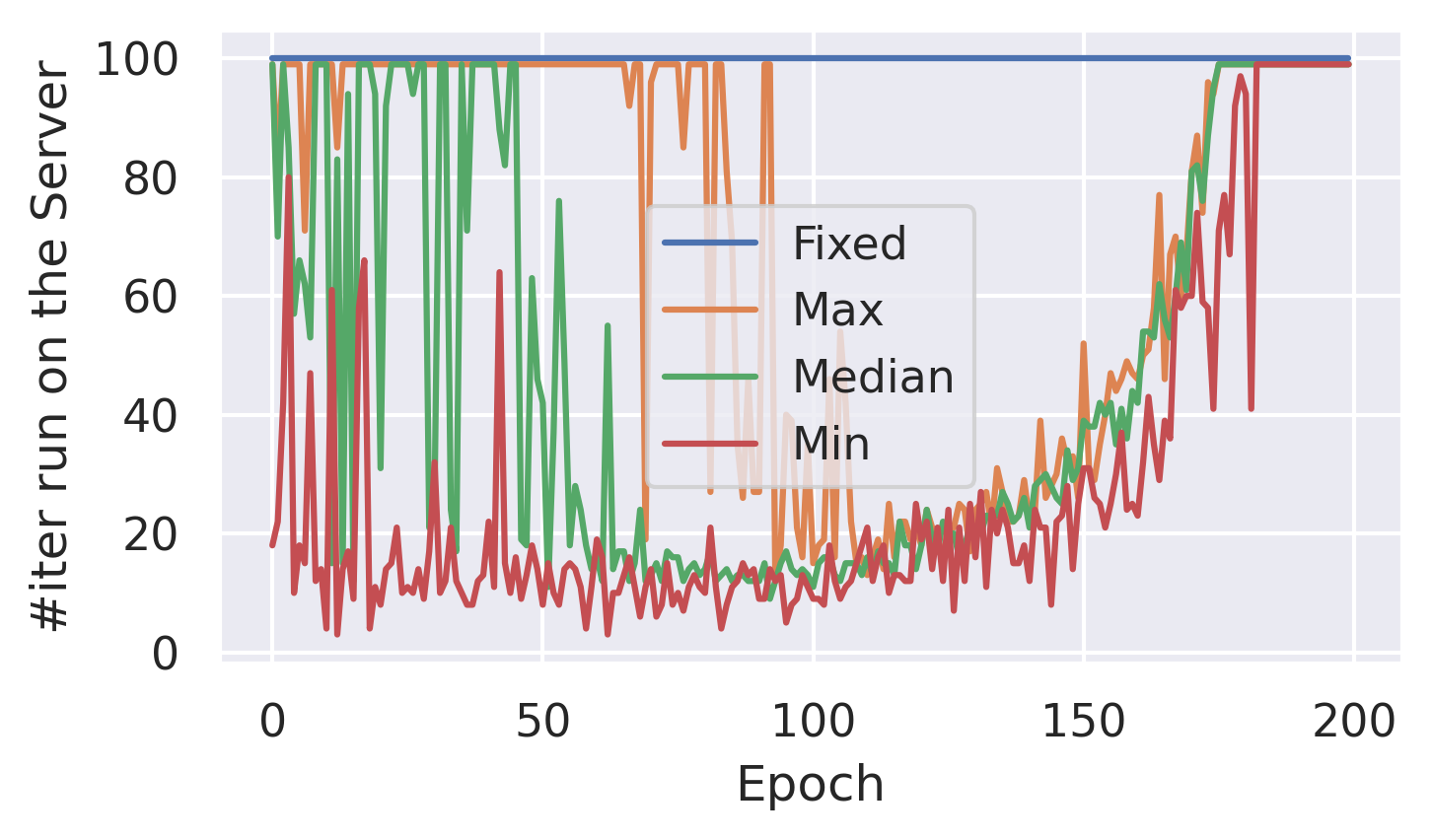}
    \caption{Training iterations for different radius selection strategies.}
    \label{fig:appendix_strategy_iteration}
\end{figure}

\subsection{Radius Selection}
\label{ssec:appendix-radius-selection}
\begin{figure}[tb]
    \centering
    \includegraphics[width=\linewidth]{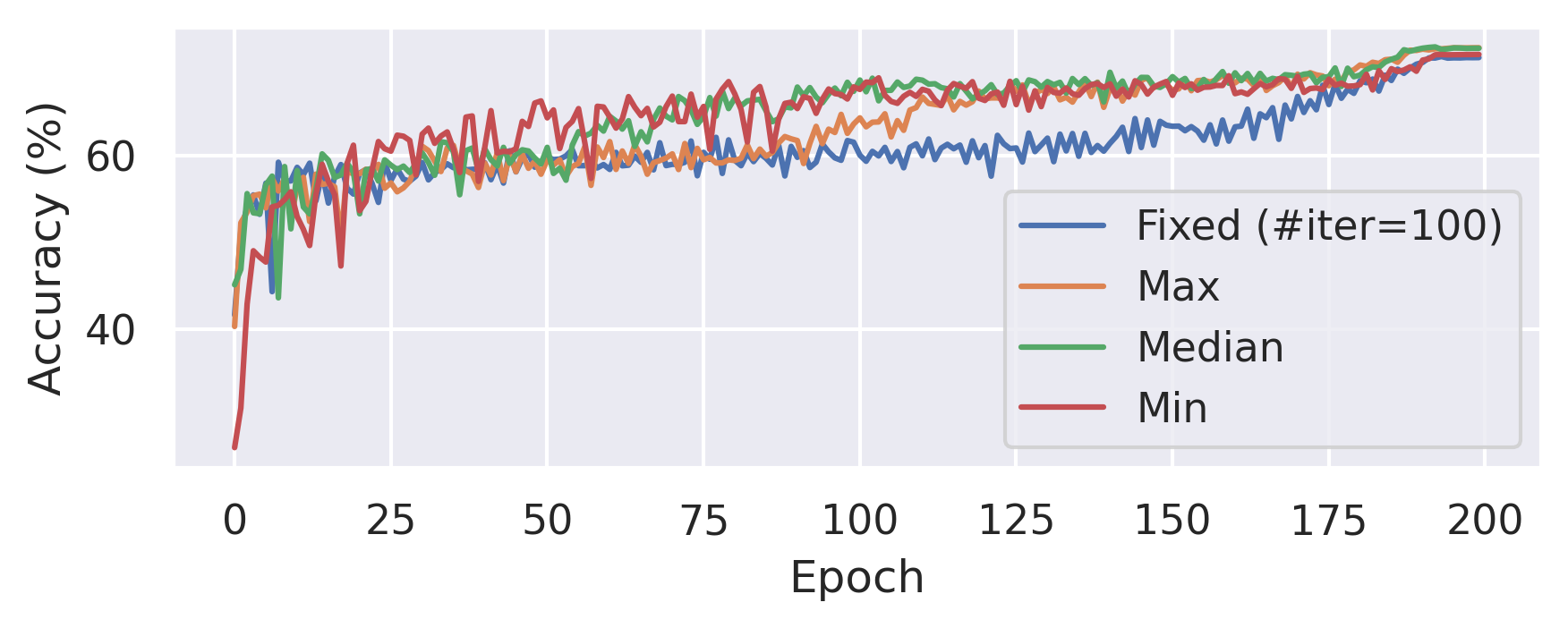}
    \caption{Ablation study on radius selection.}
    \label{fig:strategy}
\end{figure}
\begin{figure}[tb]
    \centering
    \includegraphics[width=\linewidth]{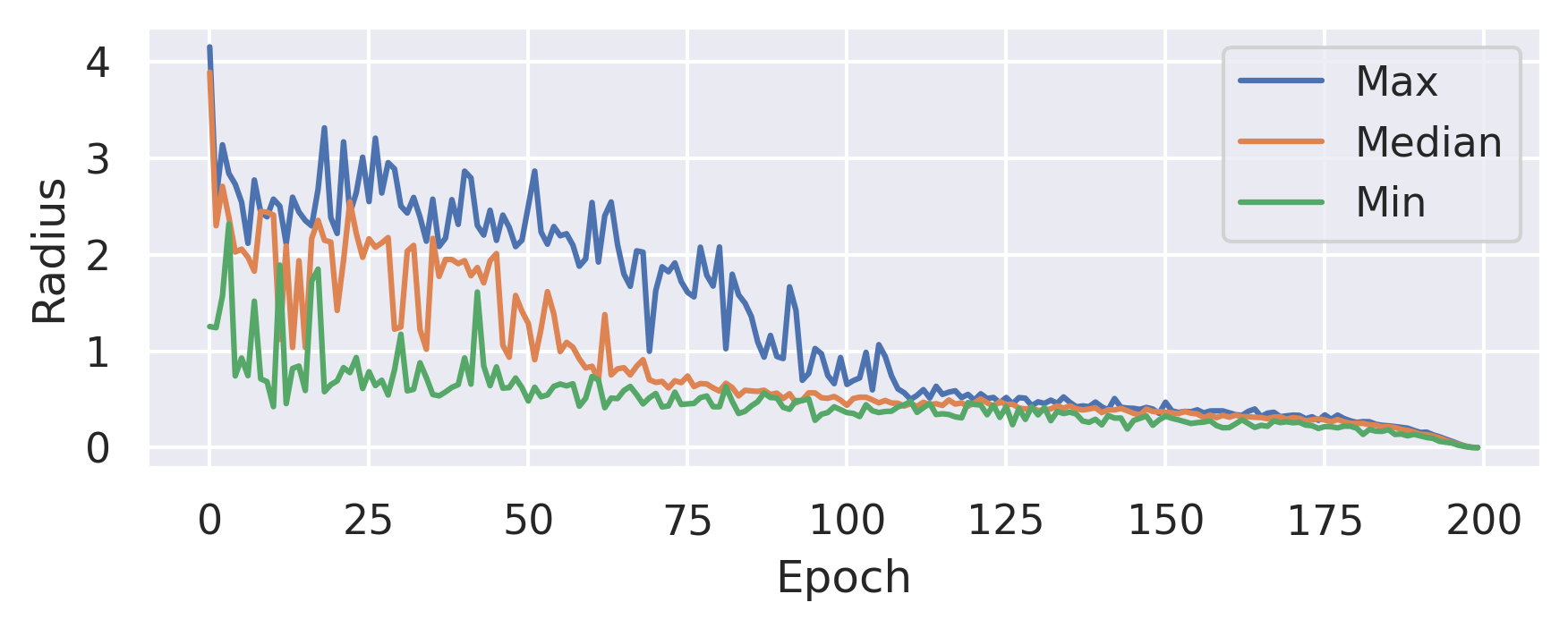}
        \caption{Radius suggested by different strategies.}
        \label{fig:strategy_radius}
\end{figure}

\begin{table}
\aboverulesep=0.1ex
\belowrulesep=0.1ex
    \begin{tabularx}{\linewidth}{@{}YYYY@{}}
        \toprule
        Min & Max & Median & Fixed \\ \midrule
        71.90 & 72.39 & 72.33 & 71.26 \\ \bottomrule
    \end{tabularx}
    \caption{Performance comparison between radius selection strategies. }
    \label{table:appendix-strategy}
\end{table}

As in \sectionautorefname~\ref{sec:method}, we evaluate different radius selection strategies for server optimization. We consider four strategies: \emph{Fixed}, \emph{Max}, \emph{Median}, and \emph{Min}. \emph{Fixed} uses a fixed length of 100 iterations, ignoring the quality. \emph{Max} aims for the fastest optimization by using the largest radius. \emph{Median} optimizes in a moderate way by considering the majority. \emph{Min}, adopted in all experiments, focuses on the safest region agreed upon by all synthetic image sets. \tableautorefname~\ref{table:strategy} reveals that the proposed strategies consistently outperform the fixed strategy. Meanwhile, \figureautorefname~\ref{fig:strategy} presents that a more aggressive strategy leads to worse intermediate performance, which may not be suitable for federated applications requiring satisfactory intermediate performance. Among them, \emph{Min} delivers the best results.
Finally, \figureautorefname~\ref{fig:strategy_radius} demonstrates that the radii proposed by different strategies change across epochs, indicating that a naively set training iteration may not be optimal. Additionally, this finding suggests the possibility of designing heuristic scheduling functions for adjusting the radius in a privacy-preserving way. The corresponding server training iterations can be found in the appendix.

In \sectionautorefname~\ref{ssec:ablation}, we show that the effective approximation regions change across rounds. A fixed pre-defined training iterations may cause sub-optimal performance. To complement the experiments, we additionally plot the corresponding training iterations on the server side in \figureautorefname~\ref{fig:appendix_strategy_iteration}. We observed that \emph{Max} and \emph{Median} tend to be more aggressive by updating for more epochs, granting faster improvement, while \emph{Min} optimizes more conservatively. Interestingly, we found that all three proposed strategies exhibit similar behavior in the later stages of training, which is in stark contrast to the fixed strategy.

\begin{figure*}[tb]
    \centering
    \includegraphics[width=\linewidth]{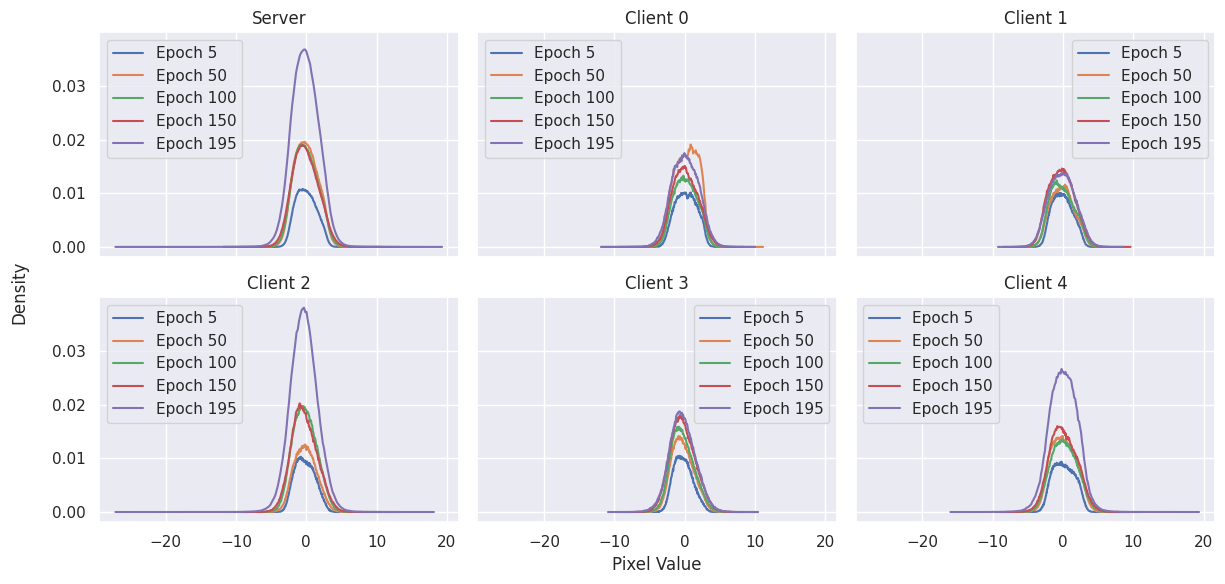}
    \caption{Pixel distributions in non-private settings. The plot illustrates the evolution of pixel values in synthetic images during training. At the early training stage, the pixel range is wider and gradually concentrates around zero as the model approaches convergence, implying that the synthetic images reflect the training status.}
    \label{fig:appendix-pixel-dist-overview}
\end{figure*}

\subsection{Qualitative Results}
\label{ssec:qualitative_result}

In this section, we examine the synthetic images. In non-private settings, \figureautorefname~\ref{fig:appendix-pixel-dist-overview} displays the pixel value distributions of all synthetic images at epochs 5, 50, 100, 150, and 195, both on the server and the clients. We made several observations from these visualizations. Firstly, within the same round, the distributions of pixel values on each client exhibit distinct behavior, indicating the diversity of private data statistics across clients. Secondly, the distributions also vary across different epochs. At earlier stages, the synthetic images present a wider range of values, gradually concentrating around zero as training progresses. This phenomenon introduces more detailed information for training and demonstrates that our method faithfully reflects the training status of each client. 
Additionally, \figureautorefname~\ref{fig:vis} displays visual examples of synthetic images corresponding to the labels "airplane," "automobile," "bird," and "cat." These visuals indicate that earlier epoch images exhibit larger pixel values and progressively integrate more noise during training, reflecting gradient details. It's crucial to underscore that our method is \emph{not intended} to produce realistic data but to approximate loss landscapes.

\section{Computation Complexity}
\label{sec:appendix-computation}
Our method suggests an alternative to current research, trading computation for improved performance and communication costs incurred by slow convergence and biased optimization. We present the computation complexity analysis for one communication round below and conclude with empirical evidence that the additional overhead is manageable, especially for cross-silo scenarios. 

We begin with the SGD complexity $\mathcal{O}(d)$, where $d$ denotes the number of network parameters. Suppose synthetic samples contain $p$ trainable parameters; then the complexity can be formulated as follows.

\begin{equation*}
\begin{split}
    &\mathcal{O}\biggl(R_i \cdot 5\cdot \Bigl(N(d + 2R_b(d+p)) + R_l d\Bigl)\biggl) \\
    =\quad &\mathcal{O}\biggl(5 R_i\cdot (2NR_b+R_l+N)d + 10R_iR_b p \biggl) \\ 
    =\quad &\mathcal{O}\biggl(5 R_iN\cdot (2R_b+\frac{R_l}{N}+1)d + 10R_iR_b p \biggl)
\end{split}    
\end{equation*}

Note that $5R_iN$ determines how much real data we will see during synthesis. For comparison, we make $5R_iN$ equal in both our method and gradient-sharing baselines (i.e., five local epochs in FedAvg with complexity $\mathcal{O}(5R_iNd)$). Overall, our method introduces $2R_b + \frac{R_l}{N}+1$ times more computation on network parameters and an additional $10R_iR_b$ term for updating synthetic samples.